\documentclass[10pt,twocolumn,letterpaper]{article}

\usepackage[pagenumbers]{cvpr}

\usepackage[T1]{fontenc}
\usepackage{cuted}
\usepackage{currfile}
\usepackage{caption}
\usepackage{url}
\usepackage{enumitem}
\usepackage{multirow}
\usepackage{amsthm}
\usepackage{algorithm}
\usepackage{algpseudocode}

\theoremstyle{remark}

\usepackage{amsmath,amsfonts,bm}

\def\eqref#1{equation~\ref{#1}}

\def\1{\bm{1}}

\def\va{{\bm{a}}}

\def\ve{{\bm{e}}}

\def\vq{{\bm{q}}}

\def\mE{{\bm{E}}}

\def\mK{{\bm{K}}}

\def\mR{{\bm{R}}}

\def\mX{{\bm{X}}}

\DeclareMathAlphabet{\mathsfit}{\encodingdefault}{\sfdefault}{m}{sl}
\SetMathAlphabet{\mathsfit}{bold}{\encodingdefault}{\sfdefault}{bx}{n}

\newcommand{\suppdoc}{\textbf{Supplementary Document}}

\newcommand{\qstar}{\vq^\star}
\newcommand{\qbar}{\bar{\vq}}

\newcommand{\ahatV}{\hat{\va}_{\mathcal V}}

\newcommand{\Ours}{ReSQue\xspace}

\usepackage{soul}
\sethlcolor{yellow!40}
\soulregister\citep7
\soulregister\citet7
\soulregister\ref7

\definecolor{cvprblue}{rgb}{0.21,0.49,0.74}
\usepackage[breaklinks,colorlinks,allcolors=cvprblue]{hyperref}

\title{\Ours: Representative Surrogate Query\\ for Streaming Video KV Cache Compression}

\author{
Daehyeon Choi$^{1}$ \quad Sangwoo Youn$^{1}$ \quad Chanho Park$^{1}$ \quad Jaihoon Kim$^{1}$\\
Phillip Y. Lee$^{1}$ \quad Byung-Kwan Lee$^{2}$ \quad Minhyuk Sung$^{1}$\\
$^{1}$KAIST \quad $^{2}$NVIDIA\\
Project page: \url{https://resque-streaming-kv.github.io/}
}

\begin{document}
\maketitle
\begin{strip}
\vspace{-4.0\baselineskip}
\centering
\includegraphics[width=\linewidth]{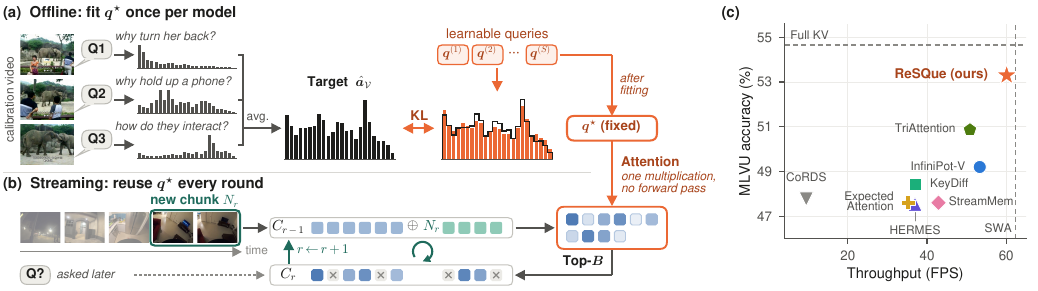}
\captionsetup{hypcap=false}
\captionof{figure}{\textbf{Overview.} (a) \Ours fits a few queries per attention head so that their attention matches the question-averaged attention on calibration videos. (b) During the stream, the fitted query $\qstar$ is reused at every compression round, scoring the cache without a forward pass. (c) MLVU accuracy at 6K against throughput, the frames processed per second, averaged over three backbones. Dashed: Full KV accuracy and SWA throughput, whose accuracy lies below the plotted range.}
\captionsetup{hypcap=true}
\label{fig:teaser}
\end{strip}

\begin{abstract}
\par
\vspace{-1.0\baselineskip}
\noindent
Streaming video understanding requires multimodal large language models (MLLMs) to compress growing KV caches before future questions are known, risking the loss of relevant information. We introduce \textbf{ReSQue} (\textbf{Representative Surrogate Query}), which directly learns reusable queries to preserve information needed by future questions. We define \emph{oracle attention} as the attention averaged over possible future questions, whose highest-scoring tokens maximize expected attention retention for a fixed cache. Since this oracle is unavailable, ReSQue constructs \emph{target attention} by averaging question attention for each calibration video and directly optimizes head-level query vectors to match its distribution. Once learned, these queries are reused across videos and compression rounds, scoring cached keys without additional model or scorer-network forward passes. We evaluate ReSQue on three MLLM backbones and four video QA benchmarks. ReSQue achieves the best or tied-best accuracy in 25 of 27 long-video settings. Further analyses demonstrate closer alignment with future-question attention and efficient streaming compression with low recurring computational overhead.
\end{abstract}

\section{Introduction}
\label{sec:intro}
\vspace{-0.5\baselineskip}

Multimodal large language models (MLLMs) are becoming a key component of everyday AI assistants, offering flexible visual understanding through natural-language interaction~\citep{bai2025qwen3vl, singh2025openai, wang2025internvl35}.
As these assistants become more widely used, their applications extend beyond images and short clips to long-form videos, including lectures, movies, egocentric recordings, and sports broadcasts.
Understanding such content requires retaining and reasoning over information that unfolds over minutes or hours, motivating recent advances in long-video understanding~\citep{song2024moviechat, he2024malmm, shen2024longvu}.

In many practical settings, however, assistants must understand video \emph{as it unfolds}, rather than after a recording is complete.
Unlike conventional long-video question answering, where the complete video and its accompanying question are provided together, \emph{streaming video understanding} requires MLLMs to process incoming video frames as they arrive, without knowing when or what questions will be asked~\citep{chen2024videollmonline, streammem2025}.
To avoid repeatedly processing past frames, MLLMs maintain a \emph{KV cache} containing the key and value vectors of previously processed visual tokens.
As new frames arrive, the cache grows continuously, making compression necessary to stay within a bounded memory budget.
Streaming introduces two challenges for this compression.
First, tokens must be selected without knowing what information future questions will require.
Second, token selection must remain efficient as the cache evolves across repeated compression rounds.
The additional computation at each round should also be limited to preserve processing throughput, measured in frames per second (FPS).

Existing approaches for streaming compression select tokens using cache statistics, as in InfiniPot-V~\citep{kim2025infinipotv}, KeyDiff~\citep{park2025keydiff}, and CoRDS~\citep{mahdizadeh2026cords}, or attention from prompt-derived \emph{surrogate queries}, as in StreamMem~\citep{streammem2025} and HERMES~\citep{zhang2026hermes}.
These surrogate queries are used to estimate which visual tokens future questions may attend to, guiding token selection during compression.
Q-Filters~\citep{godey2025qfilters} and TriAttention~\citep{mao2026triattention} leverage query statistics for accurate scoring, while Expected Attention~\citep{devoto2025expectedattention} estimates future attention under a Gaussian query distribution.
However, our analysis shows that the attention estimated by these methods often differs from that of actual video questions, and larger differences are associated with lower streaming QA accuracy (Sec.~\ref{sec:v2_mass}).
This suggests that averaging query vectors does not generally reproduce their average attention, motivating us to directly fit surrogate queries to the attention itself.
Judge Q~\citep{liu2026judgeq} and LookaheadKV~\citep{ahn2026lookaheadkv} instead learn soft input tokens to match future-response attention conditioned on an available request, while Fast KVzip~\citep{kim2026fastkvzip} learns lightweight predictors of reconstruction-based importance.
However, their learned tokens or scorers must be evaluated on each new context, introducing additional computation that can reduce processing FPS.
These approaches motivate our key question:
\textbf{Is recomputing surrogate queries from the evolving cache necessary for accurate streaming compression, or can directly learned, reusable queries provide comparable or better accuracy at lower cost?}

To better approximate the attention of future video questions while preserving processing throughput, we introduce \textbf{ReSQue} (\textbf{Representative Surrogate Query}), which learns reusable head-level queries for streaming KV cache compression.
If future questions were known, we could score each cached visual token by the average attention it receives from those questions.
We call this ideal distribution \emph{oracle attention}.
For a single selection from a fixed cache, retaining its highest-scoring tokens maximizes the attention mass preserved on average across future questions.
Since future questions are unknown during streaming, we instead construct \emph{target attention} for each calibration video by averaging the attention maps of its available questions.
ReSQue directly optimizes a small set of query vectors per attention head to match these video-specific targets.
Unlike statistical query proxies, these queries are explicitly fitted to video-question attention rather than derived solely from query statistics.
Once learned, they can be reused across videos and compression rounds to score the evolving KV cache without an additional model or scorer-network forward pass, limiting the reduction in processing FPS.

Experiments across three MLLM backbones and both long-video and streaming question-answering benchmarks demonstrate a favorable \emph{accuracy--throughput trade-off}.
ReSQue achieves the best or tied-best accuracy in 25 of 27 long-video settings.
In controlled query comparisons on held-out MLVU videos~\citep{mlvu}, ReSQue's attention more closely matches the \emph{oracle attention} than the evaluated prompt- and statistics-based proxies.
When fitted to the same target with the same attention-matching objective, directly learned queries achieve higher QA accuracy and processing FPS than soft-token and neural-scoring alternatives.
Separately, controlled comparisons show that recomputing prompt queries at every compression round provides no consistent accuracy advantage over precomputing them once.
On a single RTX PRO 6000 GPU, ReSQue sustains 58+ FPS on every backbone, about 96\% of the throughput of a sliding-window cache (SWA)~\citep{xiao2024efficient} on average at the same budget (Fig.~\ref{fig:teaser}c), while improving overall accuracy over SWA by 2.7 to 11.1 points at 6K.
These results demonstrate that ReSQue improves information preservation while largely maintaining the processing throughput of simple sliding-window caching.

\section{Related Work}
\label{sec:related_work}
\vspace{-0.3\baselineskip}

\subsection{Streaming Video Understanding}
\vspace{-0.3\baselineskip}
Prior work on streaming video understanding has explored training models for streaming inputs~\citep{chen2024videollmonline,huang2024videochatonline,wang2025streambridge} and interaction policies~\citep{qian2025dispider,ding2025streammind}, external memories and retrieval for accessing past video content~\citep{song2024moviechat,di2025rekv,ning2026livevlm,xiao2026mukv,hao2026shallowstream,zhang2024flashvstream,qian2024videostreaming,zeng2025streamforest,he2024malmm,wang2024videollamb}, and techniques for reducing visual processing costs~\citep{yao2025timechatonline,chen2025streamingtom,xie2026fluxmem}.
Under bounded memory, sliding-window approaches retain recent tokens~\citep{xiao2024efficient,xu2026streamingvlm,shen2026simplestream}, while other methods selectively preserve information from earlier frames~\citep{wu2026savemem,minh2026protokv,kwon2026deltas,song2026causalmem,ge2026selectstream}
.
In particular, InfiniPot-V~\citep{kim2025infinipotv} and CoRDS~\citep{mahdizadeh2026cords} select cached tokens based on redundancy and diversity, whereas StreamMem~\citep{streammem2025} and HERMES~\citep{zhang2026hermes} use attention from text stand-in queries to guide compression.
ReSQue repeatedly selects visual tokens within a fixed KV-cache budget for a frozen MLLM, preserving information for questions that may arrive later in the stream.

\subsection{KV Cache Compression}
\vspace{-0.3\baselineskip}
KV cache compression reduces memory via quantization~\citep{yuan2024kivi,hooper2024kvquant}, low-rank projection~\citep{chang2024palu}, merging~\citep{nawrot2024dmc,wan2024d2o}, or token eviction~\citep{zhang2024h2o}.
Most eviction methods score tokens using attention from an observation window, typically the final request-containing prompt tokens~\citep{liu2024scissorhands,li2024snapkv,cai2024pyramidkv,feng2024adakv,fu2024headkv}, or modality-aware variants~\citep{wan2024lookm,tu2025vlcache}.
In streaming, the request is unavailable, leaving only recent visual queries whose attention differs from that of future questions (Sec.~\ref{sec:v2_analysis}).

Query-agnostic methods instead use cache statistics~\citep{devoto2024knorm,park2025keydiff,chari2025compactor} or anticipate future attention from query statistics.
Q-Filters~\citep{godey2025qfilters} and TriAttention~\citep{mao2026triattention} score keys using calibrated query summaries: a principal direction or a center with per-band norms.
However, maximizing expected attention retention requires ranking by question-averaged attention (Sec.~\ref{sec:v2_target}), which is not generally determined by simple query statistics due to softmax nonlinearity.
Expected Attention~\citep{devoto2025expectedattention} estimates this average with a Gaussian query model, but relies on recent context queries, inheriting the streaming limitation.

Learned methods fit scores to future attention using request-conditioned soft input tokens~\citep{liu2026judgeq,ahn2026lookaheadkv} or neural importance predictors~\citep{moschella2026kvp,kim2026fastkvzip}, but require MLLM or predictor evaluation on each new cache.
ReSQue combines both advantages: it learns to match question-averaged attention while using fixed head-level surrogate queries that score cached keys via a single matrix multiplication, without per-round model evaluation.

\subsection{Attention as a Learning Objective}
\vspace{-0.3\baselineskip}
Attention has been widely used to distill Transformer models~\citep{jiao2019tinybert,wang2020minilm}, train retrievers from reader attention~\citep{izacard2020readerretriever}, and learn memory representations for streaming video generation~\citep{gao2026attnforcing}.
For KV cache compression, future attention provides a training signal for learned probes~\citep{liu2026judgeq,ahn2026lookaheadkv} and eviction policies~\citep{dong2026foresightkv,hauzenberger2026kvpop,moschella2026kvp}.
Building on these approaches, ReSQue uses question-averaged attention from calibration videos as a learning target, directly optimizing reusable head-level queries to preserve information relevant to unknown future questions.

\section{Preliminaries}
\label{sec:background}
\vspace{-0.3\baselineskip}
We first introduce the key concepts and notation for streaming video understanding and attention-based KV cache compression, which form the basis of our method in Sec.~\ref{sec:method}.

\vspace{-0.5\baselineskip}
\paragraph{Streaming video understanding.}
Given a video $\mathcal V$ and questions  $\mathcal U=(U_1,\dots,U_j,\dots)$ that can arrive at any time, streaming video understanding~\citep{zhang2024flashvstream,di2025rekv} aims to answer each question based on the past frames.
An MLLM's vision encoder converts each incoming frame into visual tokens.
Likewise, each question $U_j$ is represented by text tokens $\mathbf{u}_j=[u_j^{(1)},\dots,u_j^{(M)}]$, where $M$ is the number of tokens per question.

To store information from previously processed frames for future questions, the MLLM caches the keys and values computed from visual tokens in a \emph{KV cache}.
However, as streaming video frames arrive indefinitely, the MLLM must continuously compress the KV cache, risking the loss of information needed to answer the unknown future questions.
Therefore, the key challenge is to decide which visual tokens to keep while minimizing the drop in accuracy drops as little as possible.

\vspace{-0.5\baselineskip}
\paragraph{Attention-based KV cache compression.}
The MLLM processes incoming frames in chunks, with each chunk-processing step called a \emph{round}~\citep{kim2025infinipotv}.
At each round, if the cache exceeds the memory budget, every attention head of every layer compresses its own cache independently.
For a single head, let $\mathcal C$ be the set of its cached visual tokens and $\mK\in\mathbb R^{|\mathcal C|\times d}$ their cached keys (with RoPE~\citep{su2024roformer}); the compressor scores the tokens in $\mathcal C$ and keeps the $B$ highest-scoring tokens $\mathcal B\subseteq\mathcal C$, where $B$ is the budget.

Ideally, when a target question is given for the video $\mathcal V$, the queries produced by its text tokens can be used to compute attention scores for compression.
Let $\vq_j^{(i)}\in\mathbb R^d$ denote the query that the $i$-th text token of question $U_j$ produces in this head.
We then compute the attention map $ \mathcal A_{\mathcal V}$ by averaging over the $M$ queries from the question:
\vspace{-0.2\baselineskip}
\begin{equation}
  \mathcal A_{\mathcal V}(U_j)
  =\frac{1}{M}\sum_{i=1}^{M}
  \operatorname{softmax}\!\big(\vq_j^{(i)}\mK^\top/\sqrt d\big)
  \in\mathbb R^{|\mathcal C|}.
  \label{eq:question_attn}
\end{equation}
\vspace{-0.2\baselineskip}
Each entry is the average attention weight assigned to a cached visual token by the question.
The compressor uses these weights to select:
\vspace{-0.2\baselineskip}
\begin{equation}
  \mathcal B=\operatorname{Top}_{B}(\mathcal C;\va^*),
  \qquad
  \va^*=\mathcal A_{\mathcal V}(U_j),
  \label{eq:v2_single_selection}
\end{equation}
\vspace{-0.2\baselineskip}
where $\operatorname{Top}_{B}$ selects $B$ tokens from $\mathcal C$ with the highest attention weights in $\va^*$.
The keys and values of the selected tokens remain in the cache, while the rest are evicted to free memory.
By preserving tokens with high attention weights, this compression can maintain information relevant to the target question and limit the drop in answer accuracy.

In practice, future questions are unknown at compression time, ruling out direct use of the attention scores in Eq.~~\ref{eq:v2_single_selection}.
In the following section, we introduce \Ours{}, which learns a \emph{surrogate query} to approximate these scores without access to future questions.

\section{Representative Surrogate Query: \Ours{}}
\label{sec:method}
\vspace{-0.3\baselineskip}
We first identify which visual tokens should be kept to support unknown future questions (Sec.~\ref{sec:v2_target}).
This establishes a target attention distribution, which we approximate using calibration videos and questions (Sec.~\ref{sec:adm}).
We then examine alternative parameterizations (Sec.~\ref{sec:parameterization}) and introduce \Ours{}, which directly learns reusable query vectors. 

\subsection{Target Attention for Cache Compression}
\label{sec:v2_target}
\vspace{-0.3\baselineskip}
The attention-based criterion for a single question in Eq.~\ref{eq:v2_single_selection} extends to the set of future questions~$\mathcal U_{\mathcal V}$ that the same video $\mathcal V$ may receive:
\begin{equation}
  \mathcal B=\operatorname{Top}_{B}(\mathcal C;\va^*),
  \qquad
  \va^*=\mathbb E_{U\sim\mathcal U_{\mathcal V}}
  \left[\mathcal A_{\mathcal V}(U)\right].
  \label{eq:v2_expected}
\end{equation}
Each entry of $\va^*$ is the expected attention weight assigned to one cached visual token across future questions.
For a fixed cache, selecting the highest-scoring tokens maximizes the \emph{expected attention retention}, since the expected total attention weight of a selected set is the sum of $\va^*$ over it.
In streaming, however, the cache changes as new tokens arrive and earlier ones are evicted irreversibly, so this criterion is no longer guaranteed to be optimal when applied at every round.
Empirically, however, this drift does not matter: adapting the query to the video or to the stream gives no meaningful gain over a single fixed query (\S\ref{sec:v2_analysis}).

Since $\operatorname{Top}_{B}$ depends only on the ranking of the scores, selection is optimal for this objective when the scores rank tokens as $\va^*$ does, motivating attention distribution matching.
However, computing $\va^*$ requires knowledge of the future-question distribution, which is unavailable at compression time.
We therefore approximate it using calibration data, as described next.

\subsection{Attention Distribution Matching}
\label{sec:adm}
\vspace{-0.3\baselineskip}
To approximate the ideal attention score $\va^*$ in Eq.~\ref{eq:v2_expected} without access to future questions, we use a calibration dataset $\hat{\mathcal D}$ of videos $\mathcal V$ paired with finite sets of questions $\mathcal U_{\mathcal V}$ (implementation details in \suppdoc).
Attention is a distribution over the cached visual tokens of one particular video, so the target is estimated per video by averaging the attention of its calibration questions:
\begin{equation}
  \hat{\va}_{\mathcal V}
  =
  \mathbb E_{U\sim\mathcal U_{\mathcal V}}
  \left[\mathcal A_{\mathcal V}(U)\right],
  \qquad
  (\mathcal V,\mathcal U_{\mathcal V})\in\hat{\mathcal D},
  \label{eq:v2_target_estimate}
\end{equation}
where $\hat{\va}_{\mathcal V}$ estimates the target $\va^*$ of Eq.~\ref{eq:v2_expected} for the video $\mathcal V$.

We use $\hat{\va}_{\mathcal V}$ as the supervision target for learning attention scores $\va_\theta(\mathcal V)$, which are computed over the cached tokens of $\mathcal V$ with parameters $\theta$ shared across videos.
We introduce the \emph{attention distribution matching} objective, which minimizes the divergence $\operatorname{Div}$ between the target and predicted attention distributions, averaged over the calibration videos:
\begin{equation}
  \theta^\star
  =
  \arg\min_\theta\;
  \mathbb E_{(\mathcal V,\mathcal U_{\mathcal V})\sim\hat{\mathcal D}}
  \Big[
    \operatorname{Div}\big(\hat{\va}_{\mathcal V}\,\|\,\va_\theta(\mathcal V)\big)
  \Big].
  \label{eq:v2_loss}
\end{equation}
Once trained, $\va_{\theta^\star}$ provides attention scores for compressing the KV cache by selecting the top-$B$ visual tokens and evicting the rest.
The behavior of this objective depends on the parameterization of $\va_\theta$, which we examine in Sec.~\ref{sec:parameterization}.
In practice, we use the forward KL divergence; other divergences are compared in \suppdoc. 

\begin{figure*}[t]
\centering
\includegraphics[width=\textwidth]{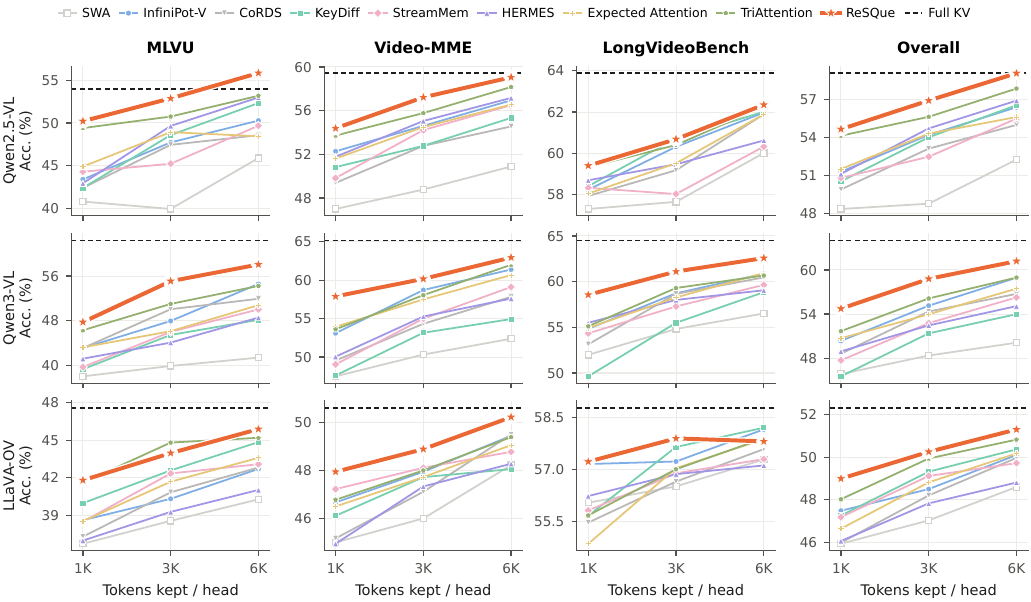}
\caption{\textbf{Long-video question answering under streaming compression.} Accuracy against the visual tokens kept per head, with the same \Ours query at every budget. Overall is the mean of the three benchmarks. Dashed: Full KV. Raw metrics reported in \suppdoc.}
\label{fig:v2_main}
\vspace{-0.75\baselineskip}
\end{figure*}

\subsection{On the Choice of Parameterization}
\label{sec:parameterization}
\vspace{-0.3\baselineskip}
The attention distribution matching objective in Eq.~\ref{eq:v2_loss} can be applied to different parameterizations of $\va_\theta$.
We compare soft tokens, neural scorers, and directly learned query under the same target with the same objective. 

\vspace{-0.75\baselineskip}
\paragraph{Neural scorers~\citep{moschella2026kvp, kim2026fastkvzip}.}
A straightforward approach is to parameterize token scores with a network $f_\theta$ that takes token representations $\mX$ as input and outputs a vector of token scores, followed by softmax to produce normalized
weights:
\begin{equation}
  \va_\theta
  = \operatorname{softmax}\!\big(f_\theta(\mX)\big),
\end{equation}
where $\mX$ consists of key and value vectors together with token positions in KVP~\citep{moschella2026kvp}, and the input hidden states of each attention layer in Fast KVzip~\citep{kim2026fastkvzip}. 
While this parameterization directly learns token scores, evaluating the network on incoming tokens introduces additional computation during streaming inference. 

\vspace{-0.75\baselineskip}
\paragraph{Soft tokens.}
Alternatively, as in Judge Q~\citep{liu2026judgeq} and LookaheadKV~\citep{ahn2026lookaheadkv}, one can parameterize $\va_\theta$ through $n$ learnable input embeddings $\mE_\theta=[\ve_\theta^{(1)},\dots,\ve_\theta^{(n)}]$, which are appended to the cache $\mathcal C$ and mapped to queries by the frozen MLLM $\Pi$; $\Pi(\mE_\theta;\mathcal C)^{(i)}$ is the query that the $i$-th embedding produces in the head when it attends to $\mathcal C$:
\begin{equation}
  \va_\theta
  =
  \frac{1}{n}\sum_{i=1}^{n}
  \operatorname{softmax}\!\big(
    \Pi(\mE_\theta;\mathcal C)^{(i)}\mK^\top/\sqrt d
  \big).
\end{equation}
During training, only the input embeddings are updated while the MLLM parameters remain fixed.
However, the queries depend on the cache they are appended to, so they must be recomputed through the MLLM at every compression round, increasing the computational cost.

\vspace{-0.75\baselineskip}
\paragraph{\Ours{}: query vectors as parameters.}
We introduce \Ours{}, which learns the query vectors themselves as parameters, so that neither a scorer network nor an MLLM forward pass is needed to obtain them. 
For each attention head, we optimize $S$ query vectors
$\vq_\theta^{(1)},\dots,\vq_\theta^{(S)}$.
Since cached keys carry rotary positional embeddings (RoPE)~\citep{su2024roformer} while these queries do not, we apply the rotary matrix $\mR_p$ to each query, where $p$ is the position at which a question would be appended; this holds in both training and inference.
The queries produce:
\begin{equation}
  \va_\theta
  =
  \frac{1}{S}\sum_{s=1}^{S}
  \operatorname{softmax}\!\big(
    \vq_\theta^{(s)}\mR_p\mK^\top/\sqrt d
  \big).
  \label{eq:v2_rows}
\end{equation}
We fit these query vectors using the attention distribution matching objective in Eq.~\ref{eq:v2_loss}. 
Since these queries are fixed and do not depend on the cache, scoring is one batched matrix multiplication between the query vectors and the cached keys, which takes about $0.01$\,s, a tenth of a compression round (Fig.~\ref{fig:v2_efficiency}a). In Sec.~\ref{sec:v2_analysis} (Fig.~\ref{fig:v2_param_main}), we show that directly learned queries achieve the highest accuracy on all three backbones, while soft tokens and neural scorers run 1.3--1.4$\times$ and 2.1--5.2$\times$ slower, respectively.

\section{Experiments}
\label{sec:v2_experiments}
\vspace{-0.3\baselineskip}
We evaluate \Ours{} as a streaming compressor against prior compressors (Sec.~\ref{sec:v2_main}) and measure its cost (Sec.~\ref{sec:v2_efficiency}); Sec.~\ref{sec:v2_analysis} then examines the query itself.

\subsection{Setup}
\label{sec:v2_setup}
\vspace{-0.3\baselineskip}

\paragraph{Models and benchmarks.}
We use Qwen2.5-VL-7B-Instruct~\citep{bai2025qwen25vl}, Qwen3-VL-8B-Instruct~\citep{bai2025qwen3vl}, and LLaVA-OneVision-7B~\citep{li2025llava_onevision}.
We evaluate on three long-video question-answering benchmarks, processed as streams: MLVU~\citep{mlvu} (502 multiple-choice questions, accuracy averaged over its holistic, single-detail and multi-detail task groups), Video-MME~\citep{fu2025videomme} (1,800 questions on its medium and long videos) and LongVideoBench~\citep{wu2024longvideobench} (validation set, 1{,}337 questions).
We also evaluate on a streaming benchmark whose questions are asked during the stream, VStream-QA~\citep{zhang2024flashvstream} (its ego-centric and movie splits, RVS-Ego and RVS-Movie, total 121 videos and 3,370 questions), whose open-ended answers are scored by an LLM judge.

\vspace{-0.75\baselineskip}
\paragraph{Streaming protocol.}
The video is processed in chunks of 6 frames, and after each chunk the compressor reduces the cached visual tokens to a fixed budget before any question is known.
On the three long-video benchmarks we test budgets of about 6K, 3K and 1K visual tokens per attention head, and on VStream-QA a budget of 6K.

\begin{figure*}[t]
\centering
\includegraphics[width=\textwidth]{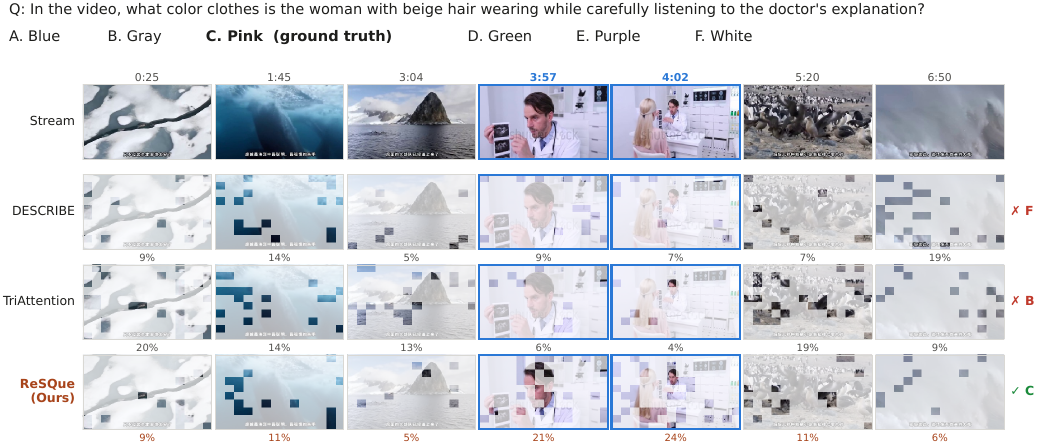}
\caption{\textbf{Qualitative example} (MLVU, Qwen3-VL, 6K tokens per head). Patches each query keeps in one head (layer 26, KV head 6) when the question arrives, with evicted patches faded and the share kept under each frame. The blue frames lie in the evidence clip (3:55--4:05), which receives 46\% of the question's attention on the uncompressed cache but spans only 2.5\% of the frames. \Ours keeps the most of the evidence frames and is the only one to answer correctly.}
\label{fig:qual}
\vspace{-\baselineskip}
\end{figure*}

\vspace{-0.75\baselineskip}
\paragraph{Baselines.}
We compare with query-free compressors, InfiniPot-V~\citep{kim2025infinipotv}, CoRDS~\citep{mahdizadeh2026cords}, and KeyDiff~\citep{park2025keydiff}; attention-based compressors that score with a text stand-in, StreamMem~\citep{streammem2025} and HERMES~\citep{zhang2026hermes}; and query-statistics-based compressors, Expected Attention~\citep{devoto2025expectedattention} and TriAttention~\citep{mao2026triattention}. We also report a sliding window that keeps the system prompt and the most recent frames (SWA), as in StreamingLLM~\citep{xiao2024efficient}.
We use the official configuration for each baseline, one setting per backbone for all benchmarks and budgets, except TriAttention, whose query statistics we compute on the NExT-QA questions that \Ours is fitted on. 

\vspace{-0.75\baselineskip}
\paragraph{Implementation details.}
\label{sec:v2_impl}
We fit \Ours (Eq.~\ref{eq:v2_loss}, with $\operatorname{Div}$ the KL divergence) on the $570$ validation videos of NExT-QA~\citep{xiao2021nextqa}, which share no video or question with the evaluation benchmarks, with $S=8$ vectors per head (Eq.~\ref{eq:v2_rows}), using Muon~\citep{jordan2024muon} at a step size of $0.03$ to $0.1$.
The fit takes under two hours on a single GPU (details in \suppdoc).

\subsection{Main Results}
\label{sec:v2_main}
\vspace{-0.3\baselineskip}

\paragraph{Offline long-video benchmarks.}
On the long-video benchmarks, where the question is asked after the stream has ended, \Ours{} reaches the highest accuracy in 25 of the 27 settings across three backbones, three benchmarks and three budgets, two of them ties, using the same query at every budget with raw metrics reported in \suppdoc.
In overall accuracy, the mean over the three benchmarks, the best baseline is TriAttention at every budget and backbone.
\Ours{} leads it by $1.2$, $1.3$ and $0.5$ points at 6K, 3K and 1K on Qwen2.5-VL, by $2.2$, $2.7$ and $3.0$ on Qwen3-VL, and by $0.5$, $0.4$ and $1.0$ on LLaVA-OV.
On Qwen3-VL the margin grows as the budget shrinks, and on LLaVA-OV it is largest at 1K, when it matters most which few tokens the cache keeps.
At 6K, its overall accuracy matches that of the uncompressed model on Qwen2.5-VL and is within $1.0$ and $2.8$ points of it on LLaVA-OV and Qwen3-VL.

TriAttention, the strongest baseline, scores keys using a trigonometric series constructed from the mean pre-RoPE query over future positional offsets, together with a query/key norm term. We compute its query statistics on the same NExT-QA questions as \Ours, but TriAttention estimates importance from query statistics rather than directly fitting question-averaged attention (Sec.~\ref{sec:v2_mass}).
Other baselines vary in relative performance across benchmarks and backbones, with neither query-free nor surrogate-query-based approaches consistently dominating. The sliding window (SWA) is the least accurate baseline in eight of the nine overall settings.
\vspace{-0.75\baselineskip}
\paragraph{Streaming benchmark.}

On VStream-QA, where questions are asked while the stream is still playing, \Ours{} is the most accurate compressor over all questions at the 6K budget on Qwen2.5-VL and LLaVA-OV, by $0.5$ and $2.5$ points, and ties TriAttention on Qwen3-VL (Table~\ref{tab:v2_streaming}).
The gain comes from the movie split.
Since the questions are asked in real time, recent frames matter, and SWA, which keeps only those frames, is the best or second best on the ego-centric split on every backbone, ahead of \Ours{} on Qwen2.5-VL and Qwen3-VL.

\vspace{-0.75\baselineskip}
\paragraph{Qualitative example.}
Fig.~\ref{fig:qual} shows what three surrogate queries keep in the cache of one MLVU video, in one head, when the question arrives.
\Ours{} keeps more than twice as many patches of the two evidence frames as the \textsc{describe} prompt and TriAttention, and only \Ours{} answers correctly.

\begin{table}[b]
\vspace{-\baselineskip}
\centering
\scriptsize
\setlength{\tabcolsep}{2.2pt}
\resizebox{\columnwidth}{!}{%
\begin{tabular}{lccccccccc}
\toprule
 & \multicolumn{3}{c}{Qwen2.5-VL-7B} & \multicolumn{3}{c}{Qwen3-VL-8B} & \multicolumn{3}{c}{LLaVA-OV-7B}\\
\cmidrule(lr){2-4}\cmidrule(lr){5-7}\cmidrule(lr){8-10}
\textbf{Method} & Ego & Mov. & All & Ego & Mov. & All & Ego & Mov. & All\\
\midrule
SWA & \underline{61.9} & 46.0 & 52.9 & \textbf{66.6} & 52.1 & 58.4 & \underline{61.0} & 41.4 & 49.9\\
InfiniPot-V & 60.4 & \underline{49.1} & 54.0 & 65.6 & 53.3 & 58.6 & 60.1 & 41.2 & 49.4\\
CoRDS & \textbf{62.6} & 48.1 & \underline{54.4} & 65.6 & 54.0 & 59.0 & 60.5 & 41.9 & 50.0\\
KeyDiff & 59.3 & \underline{49.1} & 53.5 & 65.1 & 50.9 & 57.1 & 59.0 & 42.5 & 49.7\\
StreamMem & 58.0 & 47.8 & 52.2 & 65.6 & 52.9 & 58.4 & 59.7 & \underline{42.6} & \underline{50.1}\\
HERMES & 59.5 & 48.2 & 53.1 & \underline{66.5} & 53.4 & \underline{59.1} & 60.7 & 41.4 & 49.8\\
Expected Attention & 60.1 & 48.5 & 53.6 & 65.7 & 53.3 & 58.7 & 59.5 & 41.9 & 49.5\\
TriAttention & 59.7 & 48.0 & 53.1 & 65.4 & \textbf{55.4} & \textbf{59.7} & 59.2 & 41.9 & 49.4\\
\midrule
\textbf{\Ours{}} & 60.4 & \textbf{50.6} & \textbf{54.9} & 66.0 & \underline{54.9} & \textbf{59.7} & \textbf{61.3} & \textbf{46.0} & \textbf{52.6}\\
\bottomrule
\end{tabular}}
\vspace{-0.3\baselineskip}
\caption{\textbf{Streaming video QA on VStream-QA}~\citep{zhang2024flashvstream}, $\sim$6K tokens/head: accuracy (\%) on its ego-centric (Ego) and movie (Mov.) splits and on all questions (All). Open-ended answers are judged correct or not by an LLM (\texttt{gpt-5-mini}). \textbf{Bold}: best; \underline{underline}: second.}
\label{tab:v2_streaming}
\vspace{-1\baselineskip}
\end{table}

\subsection{Efficiency}
\label{sec:v2_efficiency}

\begin{figure*}[t!]
\centering
\includegraphics[width=\textwidth]{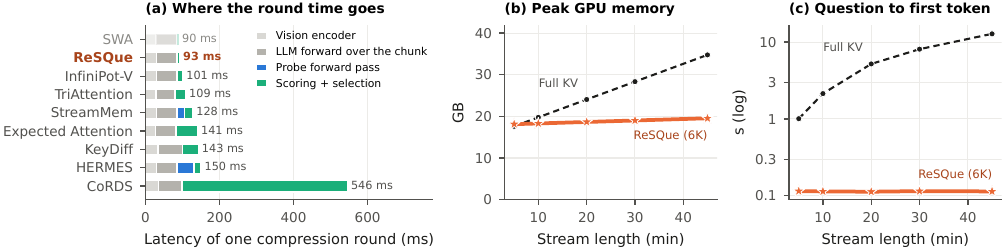}
\caption{\textbf{Cost of streaming compression} on Qwen2.5-VL (Video-MME videos of at least 45 minutes, one RTX PRO 6000, mean over 20 videos). (a) Latency of one compression round on a 6-frame chunk at 6K, averaged over the steady-state rounds. (b) Peak GPU memory and (c) time from the question to the first token on the first $T$ minutes of each video at 0.5 fps. Full KV prefills the whole stream after the question, and \Ours{} is timed for one round plus the first token. 
We present results with varying backbones and budgets in \suppdoc.}
\label{fig:v2_efficiency}
\end{figure*}

\paragraph{Compression round.}
A compression round encodes the new chunk, passes it through the model, and selects the tokens to keep.
At $93$\,ms, the round of \Ours{} is the closest to that of SWA ($90$\,ms), which keeps the most recent frames without scoring any token (Fig.~\ref{fig:v2_efficiency}a).
Encoding the chunk and the forward pass over it, which every method pays, take $84$\,ms of the round, and scoring with \Ours{} takes the remaining $8$\,ms because it needs no forward pass.
The other compressors take $101$ to $150$\,ms, including TriAttention ($109$\,ms) and Expected Attention ($141$\,ms), which also score without a forward pass, and CoRDS takes $546$\,ms.
Since every method compresses after each 6-frame chunk, a round of $t$ seconds lets it process $6/t$ frames per second.
\Ours{} can thus process $65$ frames per second on Qwen2.5-VL, close to the $67$ of SWA, against $55$ for TriAttention and $11$ for CoRDS (Fig.~\ref{fig:teaser}c averages the three backbones).
A faster round therefore lets the compressor keep up with a stream sampled at a higher frame rate.

\begin{figure*}[t!]
\centering
\includegraphics[width=\textwidth]{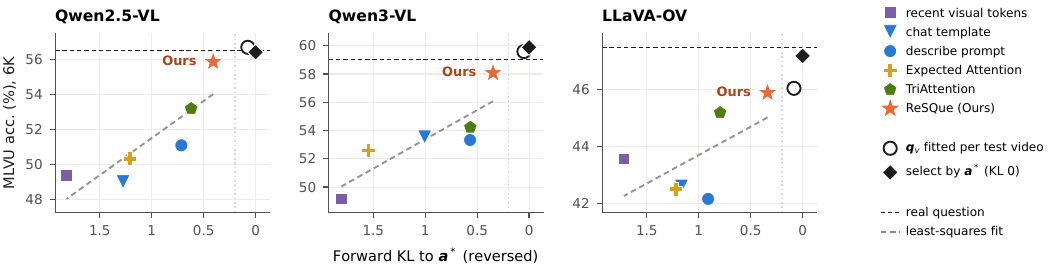}
\vspace{-1.3\baselineskip}
\caption{\textbf{Surrogate queries: closeness to the questions and accuracy.} Dashed: the real question, unknown to the compressor. Gray line: least-squares fit over the six deployable queries. Right of the dotted line: oracles that use each video's own questions.}
\label{fig:v2_gap}
\vspace{-\baselineskip}
\end{figure*}

\begin{figure}[b!]
\centering
\includegraphics[width=\columnwidth]{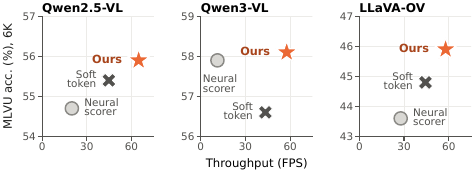}
\vspace{-1.5\baselineskip}
\caption{\textbf{Parameterization of the surrogate attention map.} MLVU accuracy at 6K against throughput, the frames processed per second (Sec.~\ref{sec:v2_efficiency}). All three are fitted with our target and objective (Eq.~\ref{eq:v2_loss}) on NExT-QA, the soft token with $n{=}32$ input embeddings.}
\label{fig:v2_param_main}
\vspace{-0.5\baselineskip}
\end{figure}

\vspace{-0.75\baselineskip}
\paragraph{Memory and time to the first token.}
Fig.~\ref{fig:v2_efficiency}b,c compares \Ours{} with Full KV when a question arrives, as the stream grows from 5 to 45 minutes.
Full KV prefills the whole stream once the question arrives, so its time to the first token grows from $1.0$ to $12.8$\,s and its peak memory from $17.6$ to $34.8$\,GB.
\Ours{} has already processed the stream, so the first token follows the question in $0.11$\,s at every length, including the round for the last chunk, and its peak memory stays between $18.1$ and $19.5$\,GB, most of it the model weights.

\section{Analysis}
\label{sec:v2_analysis}
\label{sec:v2_mass}
\vspace{-0.3\baselineskip}

In this section, we analyze why a single fitted query can stand in for questions it has never seen.
We compare \Ours with the queries of prior methods in how closely they follow the target attention and in accuracy, and with learned alternatives that compute their scores from the cache.
Further analyses of the constant query, the calibration data and the design of the query are in \suppdoc.
Unless stated otherwise, \Ours has $S=8$ vectors per head (Eq.~\ref{eq:v2_rows}), as in deployment.

\vspace{-0.75\baselineskip}
\paragraph{Setup.} We analyze on the full MLVU benchmark~\citep{mlvu}, 349 videos and 502 questions, none used for calibration, and give three numbers for Qwen2.5-VL, Qwen3-VL and LLaVA-OV, in this order.
\emph{KL} is the forward KL divergence between the target attention $\va^*$ (Eq.~\ref{eq:v2_expected}) and the attention that a query induces over the cached visual tokens, the objective of Eq.~\ref{eq:v2_loss} with $\va^*$ in place of $\ahatV$, and lower is better.
On an evaluation video, $\va^*$ averages the attention maps of its benchmark questions, an oracle that no compressor can compute, whereas $\ahatV$ is the estimate that \Ours obtains from the calibration videos.
KL is measured once on the uncompressed cache and averaged over the videos, and \emph{Acc} is the MLVU accuracy under streaming compression at the 6K budget.

\paragraph{Prior surrogate queries.}
We compare \Ours with five surrogate queries of prior methods.
Three are text queries: the \textsc{describe} instruction~\citep{kim2024infinipot,kim2025infinipotv,zhang2026hermes}, the assistant chat template~\citep{streammem2025}, and the mean query of calibration text in TriAttention~\citep{mao2026triattention}.
Two are vision queries, computed from the visual tokens in the cache: the queries of the most recent visual tokens, as in SnapKV~\citep{li2024snapkv}, and their statistics, from which Expected Attention~\citep{devoto2025expectedattention} estimates the attention of future queries.
To isolate the effect of the query itself, in Fig.~\ref{fig:v2_gap} each query only induces an attention distribution over the cached visual tokens by being scored against the cached keys, separated from the other components of its method (details in \suppdoc); the full methods, as originally proposed, are compared in Fig.~\ref{fig:v2_main}.

In Fig.~\ref{fig:v2_gap}, \Ours is the closest to the target attention and the most accurate of the deployable queries on every backbone.
TriAttention is the most accurate prior surrogate on every backbone, $2.7$, $3.8$ and $0.7$ points below \Ours.
Across the six deployable queries, Acc tends to increase as KL decreases.
This trend is strong on the Qwen models, with a Spearman correlation of $0.94$ between the negative KL and Acc, and weaker on LLaVA-OV, with $0.49$.

\vspace{-0.75\baselineskip}
\paragraph{Vision queries are far from the questions.}
Questions are text that reads the video, whereas a visual token attends to other visual tokens, so its query attends unlike a question.
Accordingly, the vision queries are farther from the target attention than the queries of words, with a KL of $1.21$ to $1.82$ against $0.57$ to $0.91$ for \textsc{describe} and TriAttention, and on the Qwen models both are less accurate than the queries of words.
A query built from the video's own tokens thus cannot replace one that knows what questions look like, which is why \Ours is fitted on video-question pairs.

\vspace{-0.5\baselineskip}
\paragraph{Directly fitting attention outperforms query statistics approaches.}
TriAttention~\citep{mao2026triattention} and Expected Attention~\citep{devoto2025expectedattention} estimate token importance from query statistics, whereas \Ours directly fits its queries to the attention induced by video questions.
In Fig.~\ref{fig:v2_gap}, both statistical approaches exhibit larger discrepancies from the distribution of real questions' attention and lower streaming accuracy than ReSQue across all three backbones.
Notably, even when TriAttention uses the same calibration questions as ReSQue, directly fitting their attention yields substantially closer alignment.

\vspace{-0.75\baselineskip}
\paragraph{Reusing learned queries improves both throughput and accuracy.}
\Ours learns query vectors once during calibration and reuses them across compression rounds, directly scoring the evolving cache without recomputing the queries.
We compare this approach with two alternatives trained using the same target attention and objective (Fig.~\ref{fig:v2_param_main}): a Judge Q-like soft-token approach~\citep{liu2026judgeq}, which generates queries through the frozen MLLM at every round, and a \emph{KVP-like neural scorer}~\citep{moschella2026kvp}, which evaluates a learned importance predictor on the current cache.
\Ours achieves $58$--$65$ FPS, running $1.3$--$1.4\times$ faster than the soft-token approach and $2.1$--$5.2\times$ faster than the neural scorer, while also improving QA accuracy by $0.5$--$1.5$ and $0.2$--$2.3$ points, respectively.
These results demonstrate that reusing directly learned queries provides both higher accuracy and greater processing throughput than the evaluated alternatives requiring additional computation at each round.

\vspace{-0.75\baselineskip}
\paragraph{Further analyses.}
Additional analyses suggest that a single reusable query can effectively serve different questions, videos, and compression rounds.
Selecting tokens using each video's question-averaged oracle attention $\va^*$ achieves accuracy within $0.9$ points of using the actual question, while fitting a separate query $\vq_{\mathcal V}$ for each video improves over \Ours by only $0.8$, $1.5$, and $0.2$ points (Fig.~\ref{fig:v2_gap}).
Similarly, recomputing prompt-derived queries at every round provides no consistent accuracy advantage over precomputing them once, with differences ranging from $-1.6$ to $+1.2$ points.
Finally, all experiments use calibration data disjoint from the evaluation benchmarks, demonstrating transfer to unseen videos and questions.
Calibration with video-question pairs outperforms text-only calibration, and even a small number of videos yields substantial gains (details in \suppdoc).

\vspace{-0.5\baselineskip}
\section{Conclusion}
\label{sec:conclusion}
\vspace{-0.5\baselineskip}
We introduced ReSQue, which directly learns reusable head-level queries to match question-averaged attention for streaming KV cache compression. Across three MLLM backbones and four video QA benchmarks, ReSQue achieves a favorable accuracy–throughput trade-off without auxiliary forward passes during compression. These results suggest that directly fitting reusable queries can provide an effective alternative to recomputing surrogate queries as the stream evolves.

\clearpage
\newpage 

\section*{LLM Usage Statement}
\vspace{-0.3\baselineskip}
We used Claude Opus 5.5, a large language model, to edit and revise the text of the paper for grammar and wording, and to revise the plotting code for the figures.
The authors reviewed and verified all text, figures and code produced with its help and take full responsibility for the content of the paper.

{
    \small
    \bibliographystyle{ieeenat_fullname}
    \bibliography{main}
}

\onecolumn

\appendix
\section*{Supplementary Material}

In this supplementary material, we first describe the implementation details of \Ours{} and the evaluation protocol with the settings of every compared method (Supp.~\ref{app:v2_impl}).
We then report additional results, the cost of \Ours{} on all backbones and the calibration statistics of TriAttention (Supp.~\ref{app:add_results}), give additional analyses, namely the full numbers of the query analysis, the KL and the attention retained by a single selection and under streaming compression (Supp.~\ref{app:v2_streaming_gap}), why a constant query suffices (Supp.~\ref{sec:v2_sensitivity}) and other parameterizations of the query (Supp.~\ref{app:v2_param_ablation}), and ablate the design choices of the query, the fit and the selection (Supp.~\ref{sec:v2_head}).
Finally, we discuss the limitations and future directions of \Ours{} (Supp.~\ref{app:limitations}) and report the full results (Supp.~\ref{app:full_results}) and additional qualitative results (Supp.~\ref{app:qualitative}).

\section{Implementation and Evaluation Details}
\label{app:v2_impl}

This section gives the implementation details of \Ours{}, which Sec.~\ref{sec:adm} defines and Sec.~\ref{sec:v2_setup} evaluates.
Algorithm~\ref{alg:v2_resque} gives the two stages: the fit of the query once per model (Supp.~\ref{app:v2_fit}) and the streaming compression with the fitted query (Supp.~\ref{app:v2_inference}).
It then describes the evaluation of Sec.~\ref{sec:v2_experiments} and Sec.~\ref{sec:v2_analysis}: the protocol shared by all methods (Supp.~\ref{app:v2_protocol}), the baselines of the main comparison (Supp.~\ref{app:v2_main_cfg}), the queries compared in Sec.~\ref{sec:v2_analysis} (Supp.~\ref{app:v2_query_cfg}) and the parameterizations compared in Sec.~\ref{sec:parameterization} (Supp.~\ref{app:v2_param_cfg}).

\subsection{Fitting the Query to the Target Attention}
\label{app:v2_fit}

\paragraph{Calibration data.}
We fit \Ours on the validation set of NExT-QA~\citep{xiao2021nextqa}, which shares no video or question with the evaluation benchmarks: its $570$ validation videos with all their questions ($5{,}343$).
As the calibration question, we use the question text only: the answer options and the answer are not used.
Frames are sampled as in the evaluation (Supp.~\ref{app:v2_baselines}).
Ablation results on the source and the amount of the calibration data are provided in Supp.~\ref{app:v2_calib}.

\paragraph{Target attention.}
Each calibration video is prefilled once into an uncompressed cache, and each of its questions is appended after the last visual token.
We write $p_t$ for the position $p$ of Eq.~\ref{eq:v2_rows} at time $t$, the position a question appended at time $t$ takes: the next M-RoPE position after the streamed tokens on Qwen2.5-VL~\citep{bai2025qwen25vl} and Qwen3-VL~\citep{bai2025qwen3vl}, with the same id in its temporal, height and width sections, and the next 1D position on LLaVA-OV~\citep{li2025llava_onevision}.
For every layer and query head, the attention of a question over the visual tokens of this cache is the average over the question's tokens of the softmax of its queries (Eq.~\ref{eq:question_attn}), and the target $\ahatV$ (Eq.~\ref{eq:v2_target_estimate}) averages it over the questions of the video.

\paragraph{Objective.}
In each head, the $S$ vectors give $\va_\theta(\mathcal V)=\frac1S\sum_{s}\operatorname{softmax}(\vq_\theta^{(s)}\mR_{p_t}\mK^\top/\sqrt d)$ over the visual tokens of $\mathcal V$ (Eq.~\ref{eq:v2_rows}).
We minimize the forward KL divergence between $\ahatV$ and $\va_\theta(\mathcal V)$ (Eq.~\ref{eq:v2_loss}).

\paragraph{Initialization and optimization.}
Every vector of a layer and head starts at $\qbar$, the mean pre-RoPE query of the calibration questions (averaged over the tokens of each question, the questions of a video and the videos), copied to each of the $S$ rows, plus independent random noise of relative norm $0.1$ for each row.
$\qbar$ is one vector per head, computed once before the fit from the same question prefill as the target attention.
We minimize the objective with Muon~\citep{jordan2024muon} optimizer, at step size $0.1$ (Qwen2.5-VL) or $0.03$ (Qwen3-VL, LLaVA-OV) for $2000$ steps with gradient clipping at $1.0$; other starting points are compared in Supp.~\ref{app:v2_init}.
Each step uses one video with all its questions; the checkpoint of the last step is used.
The fit takes $82$ / $95$ / $56$ minutes on one A6000 GPU for Qwen2.5-VL, Qwen3-VL and LLaVA-OV, and \Ours has $S$ $d$-dimensional vectors per layer and query head, $L\cdot H\cdot S\cdot d=0.8$M, $1.2$M and $0.8$M parameters.

\subsection{Inference}
\label{app:v2_inference}

The lower part of Algorithm~\ref{alg:v2_resque} gives one compression round. We write $\qstar$ for the fitted query $\vq_{\theta^\star}$ of a head, that is, its $S$ vectors $\vq_{\theta^\star}^{(s)}$.

\paragraph{One compression round.}
After a chunk is encoded, its keys and values are appended to the cache of every layer, and when the cache holds more than the budget, every layer is compressed back to $B$ visual tokens per KV head.
The system prompt is never scored or evicted, and, unlike the other compressors, \Ours{} keeps no recent window: all cached visual tokens of a head compete in one ranking, with the same budget $B$ in every KV head.

\begin{algorithm}[!ht]
\caption{\textbf{\Ours.} Fitting once per model before deployment (top) and streaming compression with the fitted query (bottom). Each query head has its own query; the query heads that share a KV head share the $B$ visual tokens it keeps.}
\label{alg:v2_resque}
\small
\textbf{Fitting} (once per model, before deployment)
\begin{algorithmic}[1]
\Require calibration videos $\hat{\mathcal D}$ with questions $\mathcal U_{\mathcal V}$; $S$ vectors per head
\State initialize every $\vq_\theta^{(s)}$ at $\qbar$ plus noise
\For{each optimization step}
  \State sample $(\mathcal V,\mathcal U_{\mathcal V})\in\hat{\mathcal D}$ and prefill $\mathcal V$ into an uncompressed cache
  \State $\ahatV\gets\mathbb E_{U\sim\mathcal U_{\mathcal V}}\big[\mathcal A_{\mathcal V}(U)\big]$ \Comment{questions appended after the last visual token; Eq.~\ref{eq:v2_target_estimate}}
  \State $\theta\gets$ one Muon step on $\operatorname{Div}\big(\ahatV\,\|\,\va_\theta(\mathcal V)\big)$ \Comment{Eq.~\ref{eq:v2_loss}}
\EndFor
\State \Return $\theta^\star\gets\theta$
\end{algorithmic}
\vspace{1pt}\hrule\vspace{3pt}
\textbf{Inference} (streaming compression)
\begin{algorithmic}[1]
\Require chunks $\mathcal N_1,\mathcal N_2,\dots$; budget $B$; fitted queries $\vq_{\theta^\star,h}^{(s)}$ of every query head $h$
\For{each chunk $\mathcal N_r$}
  \State encode $\mathcal N_r$ attending to the cache; append its key and value vectors
  \If{more than $B$ visual tokens are cached}
    \State $\va_h\gets\frac1S\sum_{s}\operatorname{softmax}\big(\vq_{\theta^\star,h}^{(s)}\mR_{p_t}\mK_{g(h)}^{\top}/\sqrt d\big)$ $\forall\,h$
    \Statex \hspace{2\algorithmicindent}$g(h)$: the KV head of query head $h$ \Comment{Eq.~\ref{eq:v2_rows}}
    \State in each KV head $g$, keep the $B$ tokens with the highest $\max_{h:\,g(h)=g}\va_h$; delete the rest \Comment{Supp.~\ref{app:gqa}}
  \EndIf
  \State \textbf{if} a question arrives \textbf{then} answer it from the cache
\EndFor
\end{algorithmic}
\end{algorithm}

\paragraph{Position ids after eviction.}
Position ids are never renumbered: the ids of the stream are computed once, as if nothing were evicted.
Keys are cached after RoPE, so evicting a token only frees its slot, while the fitted queries are stored before RoPE and rotated to $p_t$ at every round.
A question arriving later also takes $p_t$, so the query and the question see the kept tokens at the distances of the uncompressed stream.

\paragraph{Scoring under grouped-query attention.}
In each query head, each of the $S$ vectors $\vq_{\theta^\star}^{(s)}$ gives a softmax over the cached visual tokens, and the score of a token is the average of these weights over the $S$ vectors (Eq.~\ref{eq:v2_rows}).
All three backbones use grouped-query attention, in which several query heads share one KV head and thus one cache, so their scores must be combined before selection.
A KV head keeps the $B$ tokens with the largest maximum score over the query heads that share it; replacing the maximum by the mean is compared in Supp.~\ref{app:gqa}, where the maximum is slightly more accurate.
A round thus costs $S$ dot products per query head and cached token, with no forward pass and no extra KV cache.

\paragraph{Efficient inference with a fused kernel.}
Since the query is fixed and a round needs no forward pass, the scoring is a regular operation on the cached keys, which we fuse: a straightforward implementation scores one head after another, whereas a single Triton kernel~\citep{tillet2019triton} scores all heads of a layer in parallel.
The kernel fuses the rotation of $\qstar$, its product with the keys, the softmax, and the per-token score, including the maximum over the query heads that share a KV head.
The top-$B$ tokens of all heads of the layer are then selected at once.
Since the parallel kernel selects the same tokens as the sequential implementation exactly, it changes only the latency, not the accuracy.
With the kernel, the latency of a compression round (a $6$-frame chunk at the $6$K budget) falls from $148$\,ms to $93$\,ms on Qwen2.5-VL (Fig.~\ref{fig:v2_efficiency}a), from $217$\,ms to $104$\,ms on Qwen3-VL and from $166$\,ms to $104$\,ms on LLaVA-OV (Table~\ref{tab:v2_latency}).

\subsection{Common Protocol}
\label{app:v2_protocol}
\label{app:v2_baselines}

The compressor never sees the question: it compresses the stream chunk by chunk, and a question is appended only afterwards.
All methods run in one streaming pipeline with the same prompts, frame sampling and decoding, so that only the compressor differs.
Full KV, the uncompressed reference, is a single prefill over all sampled frames.
This protocol is shared by the main comparison, the comparison of queries and the comparison of parameterizations below, and Table~\ref{tab:v2_protocol} lists its settings for each backbone.

\paragraph{Frames.}
Frame sampling and the chunk setting follow InfiniPot-V~\citep{kim2025infinipotv}: frames are sampled uniformly over the whole video at a fixed rate (Table~\ref{tab:v2_protocol}), and if this yields more frames than the maximum number each backbone supports, the video is uniformly resampled to that maximum over its whole length.
Every frame is resized to at most $1280{\times}720$ before it is encoded.
The number of visual tokens per frame is kept fixed, so that a budget in frames corresponds to a fixed number of tokens.

\paragraph{Compression.}
The video is processed in chunks of $6$ frames.
Whenever the cache reaches the budget plus one chunk, it is compressed back to the budget, which corresponds to about $6$K, $3$K and $1$K visual tokens per head.
Compression applies to every layer unless noted in Table~\ref{tab:v2_baselines}, and a method that selects tokens per head keeps the same number of tokens in every head.
HERMES is the exception: it keeps the chunk size of its official code (Supp.~\ref{app:v2_main_cfg}, \textit{``Differences in cadence''}). 

\paragraph{Questions and decoding.}
On the long-video benchmarks, the question is appended after the last chunk.
On VStream-QA, the stream is processed up to the timestamp of each question: a chunk that contains the timestamp is cut at it, so that no method sees a frame after the question, and the question is appended there.
After the question is answered, the tokens of the question and its answer are removed from the cache before the stream continues, so that later chunks and questions do not see them.
When a question arrives, any cache above the budget is compressed once more, so every method answers from at most $B$ tokens per head.
Full KV caches every sampled frame and answers from the whole cache.
Answers are decoded greedily with at most $300$ new tokens.

\begin{table}[h]
\centering
\footnotesize
\setlength{\tabcolsep}{3pt}
\begin{tabular}{@{}lcccc@{}}
\toprule
Backbone & Sampling FPS & Max fr. & Tok./fr. & Budget (fr.) \\
\midrule
Qwen2.5-VL      & 2   & 768 & $\sim$140    & 42 / 21 / 7 \\
Qwen3-VL        & 2   & 768 & $\sim$140    & 42 / 21 / 7 \\
LLaVA-OV        & 0.5 & 128 & 196          & 30 / 15 / 5 \\
\bottomrule
\end{tabular}
\caption{\textbf{Streaming protocol of each backbone.} The budgets correspond to about 6K / 3K / 1K visual tokens per head, and the cache is compressed whenever it reaches the budget plus one 6-frame chunk.}
\label{tab:v2_protocol}
\end{table}

\subsection{Main Comparison}
\label{app:v2_main_cfg}

\paragraph{Baselines.}
Each baseline uses one setting per backbone for all benchmarks and budgets, taken from its public code or, where none exists, its paper.
Table~\ref{tab:v2_baselines} lists the source and the configuration of each baseline; hyperparameters it does not mention take the defaults of the source.

\begin{table*}[t]
\centering
\small
\caption{\textbf{Baseline configurations}, identical on every benchmark and budget.}
\label{tab:v2_baselines}
\begin{tabular}{@{}l p{0.17\textwidth} p{0.55\textwidth}@{}}
\toprule
Method & Source & Configuration \\
\midrule
SWA~\citep{xiao2024efficient} & reimplemented & System prompt and the most recent frames up to the budget; no scoring.  \\
InfiniPot-V~\citep{kim2025infinipotv} & \raggedright official code (\url{github.com/aiha-lab/InfiniPot-V}) & Half of the budget by the temporal-redundancy score against the most recent $25\%$ of the cached frames (always kept), the rest by value norm; with layer-wise adaptive pooling.  \\
CoRDS~\citep{mahdizadeh2026cords} & \raggedright official code (\url{github.com/ailarmhz/CoRDS}) & Coreset selection on the bottom $25\%$ of layers; the other layers reuse the indices of the nearest selected layer.  \\
KeyDiff~\citep{park2025keydiff} & \raggedright KVPress (\url{github.com/NVIDIA/KVPress}) & Negative similarity of each key to the mean key; the most recent $25\%$ of the cached frames always kept.  \\
StreamMem~\citep{streammem2025} & reimplemented & Chat-template proxy query; evicted tokens merged frame-wise by their weights, with YaRN position scaling and input-frame filtering.  \\
HERMES~\citep{zhang2026hermes} & \raggedright official code (\url{github.com/haowei-freesky/HERMES})& Shallowest $10\%$ of layers keep recent tokens, deepest $30\%$ score by attention, the rest by a `local' and a `global' probe; $16$-frame chunks; evicted tokens mean-pooled into one slot, with RoPE fold.  \\
Expected Attention~\citep{devoto2025expectedattention} & \raggedright KVPress (\url{github.com/NVIDIA/KVPress}) & Gaussian over the queries of the newest chunk, expected attention over $512$ future positions, with the value norm; $\epsilon{=}0.02$, mean over the query heads of a KV head.  \\
TriAttention~\citep{mao2026triattention} & \raggedright official code (\url{github.com/WeianMao/triattention}) & Full score, the trigonometric term plus the norm term, with query statistics computed on the NExT-QA calibration questions that \Ours{} is fitted on, namely the mean pre-RoPE query of Sec.~\ref{sec:v2_analysis} and the mean norm of each RoPE frequency pair, and future offsets $1,2,4,\dots,2^{16}$.  \\
\bottomrule
\end{tabular}
\end{table*}

\paragraph{Differences in cadence.}
HERMES keeps the $16$-frame chunks of its official code, and StreamMem, originally run on $8$-frame clips trimmed before the next clip is appended, is run on a $6$-frame cadence and trims each clip in the round it arrives, in our reimplementation.
The accuracies of HERMES use its $16$-frame chunks, whereas its round latency in Table~\ref{tab:v2_latency} and Fig.~\ref{fig:v2_efficiency} is measured with $6$-frame chunks (HERMES (6-fr.)), like every other method.

\paragraph{Offline long-video benchmarks.}
For the three long-video benchmarks, they are multiple-choice, and a prediction is correct if the first option letter in the model's answer matches the ground truth.
MLVU is scored as the mean of the accuracies of its holistic, single-detail and multi-detail task groups.
Video-MME is scored on its medium and long videos, $900$ questions each and without subtitles, as the mean of the two accuracies.
LongVideoBench is scored on its validation set as the mean over its four video-length groups ($15$, $60$, $600$ and $3600$ seconds).

\paragraph{Streaming benchmark.}
On VStream-QA~\citep{zhang2024flashvstream}, open-ended answers are judged by \texttt{gpt-5-mini} with the judging prompt and protocol of the official InfiniPot-V implementation (\texttt{rvs\_judge.py}), which follows the protocol of Flash-VStream~\cite{zhang2024flashvstream}; accuracy is the fraction of answers the judge marks as correct.

\subsection{Query Comparison}
\label{app:v2_query_cfg}

This subsection details the queries compared in Sec.~\ref{sec:v2_analysis}.
Unlike the main comparison, which runs each baseline as proposed, this comparison isolates the query: each query is only scored against the cached keys, with the selector, the streaming protocol and the budgets of Supp.~\ref{app:v2_protocol} held fixed, and every other component of the original method is left out.
The accuracy of every query at the three budgets is in Table~\ref{tab:v2_query_ext} (Supp.~\ref{app:full_results}).

\paragraph{Queries.}
Besides the \textsc{describe} instruction of the main text, we include \textsc{explain} and \textsc{summarize}, two generic prompts of the same kind, to show that the conclusions drawn from \textsc{describe} do not depend on its particular wording.
\label{app:v2_prompts}
\begin{itemize}
\item \textbf{\textsc{describe}}: \texttt{``Describe the video.''}
\item \textbf{\textsc{explain}}: \texttt{``Explain what you see.''}
\item \textbf{\textsc{summarize}}: \texttt{``Summarize the video.''}
\item \textbf{Chat template}: the generation-prompt suffix of the model's chat format, \texttt{<|im\_start|>assistant}, as in StreamMem~\citep{streammem2025}.
\item \textbf{Recent visual tokens}: the queries of the $64$ most recent visual tokens of the cache, as in SnapKV~\citep{li2024snapkv}.
\item \textbf{Q-Filters}~\citep{godey2025qfilters}: the first right singular vector of $3{,}000$ queries of the visual tokens of the NExT-QA calibration videos, per query head, averaged over the heads of a KV head, as in the original recipe; keys are scored by their inner product with this filter, without a softmax.
\item \textbf{Expected Attention}~\citep{devoto2025expectedattention}: the mean and the covariance of the queries of the visual tokens of the newest chunk, with the expected attention to each key computed as in the original method but without the value-norm factor.
\item \textbf{TriAttention}~\citep{mao2026triattention}: its full score, the trigonometric term plus the norm term. The trigonometric term is the inner product between the cached key and the mean pre-RoPE query of the NExT-QA calibration questions, rotated to future positions $1,2,4,\dots,2^{16}$ ahead of the question position; the norm term is $\sum_f(\mathbb{E}|q_f|-|\bar q_f|)\,|k_f|$ over the RoPE frequency pairs $f$.
\end{itemize}
Each text query is rotated to $p_t$ (Supp.~\ref{app:v2_fit}), and each prompt token gives one query vector.
We omit the KL of Q-Filters because it produces unnormalized importance scores rather than a native attention distribution, and report only its accuracy (Table~\ref{tab:v2_query_ext}).

\paragraph{Attention induced by each query.}
These distributions are the ones whose KL to $\va^*$ is reported in Fig.~\ref{fig:v2_gap}.
For every head, a surrogate query is a set of query vectors, each rotated to $p_t$, and it induces an attention distribution over the cached visual tokens.
\begin{itemize}
\item \textbf{\Ours}: the mean of the $S$ softmax maps of Eq.~\ref{eq:v2_rows}.
\item \textbf{\textsc{describe}, \textsc{explain}, \textsc{summarize}, chat template}: the mean of the $M$ (number of tokens) softmax maps of the query of each prompt token, as in Eq.~\ref{eq:question_attn}, computed on the uncompressed cache.
\item \textbf{Recent visual tokens}: the mean of the softmax maps of the queries of the $64$ most recent visual tokens.
\item \textbf{Expected Attention}: the softmax over cached tokens of the attention that keys receive in expectation under the Gaussian model of the queries of the visual tokens of the newest chunk.
\item \textbf{TriAttention}: the softmax over cached tokens of its full score (trigonometric term plus norm term), scaled by $1/\sqrt{d}$ like the logits of attention.
\end{itemize}

\subsection{Parameterization Comparison}
\label{app:v2_param_cfg}

This subsection details the parameterizations compared in Sec.~\ref{sec:parameterization}.
All are fitted on the $570$ NExT-QA validation videos with the target and objective of Eq.~\ref{eq:v2_loss}, and evaluated on MLVU at 6K with the selector and protocol of Supp.~\ref{app:v2_protocol}.

\paragraph{Soft tokens (Judge Q-like).}
The $n$ learned input embeddings are appended after the cache at $p_t$ and passed through the frozen model at every compression round; the query that each embedding produces in each head scores the cached keys.
Fig.~\ref{fig:v2_param_main} uses $n=32$, the default of the official Judge Q implementation~\citep{liu2026judgeq}.
The embeddings are initialized from the mean input embeddings of the NExT-QA question tokens plus noise, then trained for $2{,}000$ steps with Muon, and the final checkpoint is used.
Other numbers of embeddings are compared in Supp.~\ref{app:v2_param_ablation}.

\paragraph{Neural scorers (KVP-like).}
 One network per (layer, KV head) reads each cached visual token, namely its key, de-rotated to $p_t$, and bounded features of its position (time, row and column), and outputs one score per query head of the group.
Fig.~\ref{fig:v2_param_main} uses the MLP, with 0.85M parameters on Qwen2.5-VL.
The scorer is trained alone with Muon for $2{,}000$ steps, and the final checkpoint is used.
It reads the cached keys rather than the hidden states that Fast KVzip~\citep{kim2026fastkvzip} uses, a cache-only setting shared by all parameterizations; Supp.~\ref{app:v2_param_ablation} discusses this choice and compares architectures and inputs.

\section{Additional Results}
\label{app:add_results}
This section extends the experiments of Sec.~\ref{sec:v2_experiments} with the cost of \Ours{} on all backbones (Supp.~\ref{app:v2_efficiency}) and the calibration statistics of TriAttention in the main comparison (Supp.~\ref{app:triattention_calib}).

\subsection{Efficiency on All Backbones}
\label{app:v2_efficiency}

This subsection extends the efficiency analysis of Sec.~\ref{sec:v2_efficiency} to all three backbones.

\paragraph{Protocol.}
Table~\ref{tab:v2_latency} follows the protocol of Fig.~\ref{fig:v2_efficiency}: one RTX PRO 6000, and $22$ Video-MME videos of at least $45$ minutes, of which the first two are discarded as warm-up and the other $20$ averaged. 
A round is timed in the steady state, excluding the first and the last chunk and the rounds before the cache is full, and FPS is the $6$ frames of a chunk divided by its round time.
The answer time and the peak memory at the bottom of the table use Video-MME medium and long videos, so that all three backbones are measured on the same videos. The peak memory includes the model weights, which take 15.5, 16.3 and 15.0~GB on the three backbones, so the percentages are reductions of the total memory, not of the cache.
The answer time runs from the start of the question prefill to the end of the decoded answer, a single letter.
Full KV is the offline, memory-unconstrained setting of Table~\ref{tab:v2_main}: it prefills all visual tokens after the question, whereas \Ours{} answers from the cache it compressed during the stream.
Both are measured with every video resampled to about 50K visual tokens.

\paragraph{Results.}
A round of \Ours{} takes $0.093$ to $0.104$\,s ($58$ to $65$ FPS), within $0.006$\,s of SWA, which scores no token.
The scoring of \Ours{} is a single matrix product of its fixed query with the keys, which makes it easy to fuse with the selection into one kernel (Supp.~\ref{app:v2_inference}); the latencies of \Ours{} in (a) are measured with it, and without it a round takes $0.148$, $0.217$ and $0.166$\,s on the three backbones (Table~\ref{tab:v2_latency}(b)). TriAttention is measured in (a) with its head-batched selection. The fused selection shortens a round of \Ours{} by $37$ to $52\%$ at 6K, by $24$ to $37\%$ at 3K and by $25$ to $40\%$ at 1K, and a round of TriAttention by $14$ to $34\%$ at 6K, and by at most $26\%$ at 3K and 1K, with no gain on LLaVA-OV at 3K. In (b), the per-head rows run the selection as a loop over heads, whereas the fused rows select with the head-batched selection of TriAttention and the fused kernel of \Ours{}.
The answer time is $130$ to $170$ times shorter with \Ours{} than with offline Full KV; this compares a streaming and an offline setting, not the speed of the two per token. \Ours{} has compressed the stream before the question arrives, so only the $6$K cache is read when it is asked, whereas Full KV prefills all $50$K visual tokens after the question.
Its peak memory is $36$ to $47\%$ lower.
Most of the remaining memory is the model weights: the 6K cache takes $0.3$ to $0.8$\,GB against $2.7$ to $6.9$\,GB for 50K tokens, so lowering the budget to 1K saves little more.
The latency of CoRDS falls most with the budget ($0.546$, $0.320$ and $0.149$\,s at 6K, 3K and 1K on Qwen2.5-VL), because its selection grows with the number of kept tokens.
A round of \Ours{} takes $0.088$ to $0.101$\,s at 3K and $0.084$ to $0.097$\,s at 1K, within $0.004$\,s of SWA, which scores no token.

\begin{table*}[t]
\centering
\footnotesize
\setlength{\tabcolsep}{2pt}
\begin{tabular}{lcccccccccccc}
\toprule
method & \multicolumn{4}{c}{Qwen2.5-VL} & \multicolumn{4}{c}{Qwen3-VL} & \multicolumn{4}{c}{LLaVA-OV}\\
\cmidrule(lr){2-5}\cmidrule(lr){6-9}\cmidrule(lr){10-13}
 & 6K (s) & FPS & 3K (s) & 1K (s) & 6K (s) & FPS & 3K (s) & 1K (s) & 6K (s) & FPS & 3K (s) & 1K (s)\\
\midrule
\multicolumn{13}{l}{\emph{(a) Round latency (s) at 6K, 3K and 1K, and frames per second at 6K}}\\
\textcolor[gray]{0.35}{SWA} & \textcolor[gray]{0.35}{0.090} & \textcolor[gray]{0.35}{67.0} & \textcolor[gray]{0.35}{0.090} & \textcolor[gray]{0.35}{0.088} & \textcolor[gray]{0.35}{0.098} & \textcolor[gray]{0.35}{61.4} & \textcolor[gray]{0.35}{0.091} & \textcolor[gray]{0.35}{0.086} & \textcolor[gray]{0.35}{0.102} & \textcolor[gray]{0.35}{58.8} & \textcolor[gray]{0.35}{0.097} & \textcolor[gray]{0.35}{0.093}\\
InfiniPot-V & 0.101 & 59.3 & 0.092 & 0.089 & 0.121 & 49.5 & 0.101 & 0.095 & 0.117 & 51.1 & 0.111 & 0.105\\
KeyDiff & 0.143 & 41.9 & 0.132 & 0.129 & 0.217 & 27.7 & 0.168 & 0.162 & 0.144 & 41.8 & 0.125 & 0.127\\
CoRDS & 0.546 & 11.0 & 0.320 & 0.149 & 1.192 & 5.0 & 0.582 & 0.256 & 0.481 & 12.5 & 0.271 & 0.166\\
StreamMem & 0.128 & 46.8 & 0.135 & 0.127 & 0.154 & 38.8 & 0.151 & 0.152 & 0.139 & 43.1 & 0.129 & 0.124\\
HERMES (6-fr.) & 0.150 & 39.9 & 0.156 & 0.146 & 0.178 & 33.7 & 0.173 & 0.177 & 0.161 & 37.3 & 0.158 & 0.150\\
Expected Attention & 0.141 & 42.6 & 0.144 & 0.140 & 0.239 & 25.1 & 0.180 & 0.169 & 0.158 & 38.1 & 0.136 & 0.138\\
TriAttention & 0.109 & 55.1 & 0.108 & 0.107 & 0.122 & 49.3 & 0.111 & 0.105 & 0.125 & 48.1 & 0.116 & 0.111\\
\midrule
\textbf{\Ours{}} & \textbf{0.093} & \textbf{64.8} & \textbf{0.088} & \textbf{0.084} & \textbf{0.104} & \textbf{57.7} & \textbf{0.093} & \textbf{0.089} & \textbf{0.104} & \textbf{57.9} & \textbf{0.101} & \textbf{0.097}\\
\midrule
\multicolumn{13}{l}{\emph{(b) Round latency (s) with the per-head selection and with the fused selection, and frames per second at 6K}}\\
TriAttention (per-head) & 0.143 & 42.0 & 0.113 & 0.110 & 0.186 & 32.3 & 0.142 & 0.141 & 0.145 & 41.4 & 0.114 & 0.123\\
TriAttention (fused) & 0.109 & 55.1 & 0.108 & 0.107 & 0.122 & 49.3 & 0.111 & 0.105 & 0.125 & 48.1 & 0.116 & 0.111\\
\Ours{} (per-head) & 0.148 & 40.7 & 0.117 & 0.115 & 0.217 & 27.7 & 0.148 & 0.149 & 0.166 & 36.1 & 0.133 & 0.129\\
\Ours{} (fused) & \textbf{0.093} & \textbf{64.8} & \textbf{0.088} & \textbf{0.084} & \textbf{0.104} & \textbf{57.7} & \textbf{0.093} & \textbf{0.089} & \textbf{0.104} & \textbf{57.9} & \textbf{0.101} & \textbf{0.097}\\
\midrule
\multicolumn{13}{l}{\emph{(c) Peak memory (GB)\,/\,answer time (s): streaming (\Ours{}) vs.\ offline (Full KV)}}\\
Full KV (50K) & \multicolumn{4}{c}{26.0\,/\,6.563} & \multicolumn{4}{c}{31.0\,/\,6.340} & \multicolumn{4}{c}{29.8\,/\,4.997}\\
\Ours{} (6K) & \multicolumn{4}{c}{16.7\,/\,0.038} & \multicolumn{4}{c}{18.4\,/\,0.049} & \multicolumn{4}{c}{15.8\,/\,0.039}\\
\Ours{} (3K)  & \multicolumn{4}{c}{16.6\,/\,0.040} & \multicolumn{4}{c}{18.0\,/\,0.045} & \multicolumn{4}{c}{15.6\,/\,0.042}\\
\Ours{} (1K) & \multicolumn{4}{c}{16.5\,/\,0.035} & \multicolumn{4}{c}{17.8\,/\,0.046} & \multicolumn{4}{c}{15.5\,/\,0.035}\\
\bottomrule
\end{tabular}
\caption{\textbf{Cost of streaming compression.} (a) Latency in seconds of a 6-frame compression round at the 6K, 3K and 1K budgets and, at 6K, the FPS it allows ($=6/$latency); SWA scores no token (gray). (b) The same for TriAttention and \Ours{} with the per-head selection and with the fused selection; its fused rows are those of (a). (c) Peak memory\,/\,answer time of offline Full KV and of streaming \Ours{} at 6K, 3K and 1K; the answer times compare a streaming with an offline setting, since \Ours{} compresses the stream before the question arrives. The latencies at 3K and 1K and the memory of \Ours{} at 3K were measured separately with the same protocol.}
\label{tab:v2_latency}
\end{table*}

\begin{table}[t]
\centering
\small
\setlength{\tabcolsep}{6pt}
\begin{tabular}{lccc}
\toprule
Calibration statistics & Qwen2.5-VL & Qwen3-VL & LLaVA-OV \\
\midrule
Wikipedia text (no video) & 49.7 & 49.8 & 44.3 \\
NExT-QA questions (main comparison) & 53.2 & 54.2 & 45.2 \\
\bottomrule
\end{tabular}
\caption{\textbf{TriAttention with different calibration statistics.} MLVU accuracy (\%) at the 6K budget of TriAttention with statistics computed on Wikipedia text and on the NExT-QA questions, as in the main comparison (Table~\ref{tab:v2_main}).}
\label{tab:triattention_calib}
\end{table}
\subsection{TriAttention with Other Calibration Statistics}
\label{app:triattention_calib}

The main comparison computes the statistics of TriAttention on the NExT-QA questions that \Ours{} is fitted on (Supp.~\ref{app:v2_main_cfg}), so that the two methods see the same calibration data.
TriAttention itself calibrates on generic text, without any video or question.
This subsection shows how much that choice matters for TriAttention, with Table~\ref{tab:triattention_calib}.

\paragraph{Calibration statistics.}
In both settings the statistics are the mean pre-RoPE query of every layer and query head, and the mean norm of each RoPE frequency pair, taken over the query tokens and weighted by token; each query is de-rotated at its own position, and TriAttention has no fit.
\emph{Wikipedia text} uses generic text without any video or question, from the Wikipedia source of the calibration-source experiment (Supp.~\ref{app:v2_calib}).
\emph{NExT-QA questions} are the tokens of the questions of the NExT-QA pool of \Ours{}, computed with their video in the context.

\paragraph{The main comparison gives TriAttention its more favorable calibration.}
The statistics of the questions are more accurate than those of the text on all three backbones (Table~\ref{tab:triattention_calib}), most clearly on the two Qwen models.
With text statistics, TriAttention would fall more than twice as far below \Ours{} on every backbone, so the comparison of Table~\ref{tab:v2_main} gives TriAttention the more favorable of the two calibrations.
As for the calibration of \Ours{} (Supp.~\ref{app:v2_calib}), questions asked about videos are a better calibration source than generic text.

\section{Additional Analysis}
\label{app:add_analysis}
This section extends the analysis of Sec.~\ref{sec:v2_analysis}: the full numbers of the query analysis (Supp.~\ref{app:v2_streaming_gap}), why a constant query suffices (Supp.~\ref{sec:v2_sensitivity}), and alternative parameterizations of the query (Supp.~\ref{app:v2_param_ablation}).

\subsection{Query Analysis: KL and Attention Retention}
\label{app:v2_streaming_gap}

This subsection gives the numbers behind the query analysis of Sec.~\ref{sec:v2_analysis} and compares, for the same queries, the attention retained by a single selection on the uncompressed cache with the attention retained under streaming compression.
It then discusses which changes of the token scores between rounds can account for the difference between the two.

\paragraph{Measures.}
Table~\ref{tab:v2_query_full} reports three measures for each query.
The \emph{KL} is the forward KL divergence between the target attention $\va^*$ (Eq.~\ref{eq:v2_expected}) and the attention that a query induces over the cached visual tokens, measured once on the uncompressed cache and averaged over the 349 MLVU videos; with the accuracy, it is the content of Fig.~\ref{fig:v2_gap}, and the accuracy is that of MLVU under streaming compression at the 6K budget.
KL is measured on the uncompressed cache, whereas the compressor works on a cache that earlier rounds have thinned, so we also measure the attention that the kept tokens hold in both settings, with the \emph{expected attention retention}.
For a kept set $\mathcal B_g$ of a KV head $g$ and a question $U$ with attention $\mathcal A_g(U)$ on the uncompressed cache, normalized to sum to one over the cached visual tokens,
\[
\operatorname{ExpRet}_g(\mathcal B_g)=\mathbb{E}_U\Big[\sum_{c\in\mathcal B_g}\mathcal A_g(U)^{(c)}\Big]=\sum_{c\in\mathcal B_g}\va_g^{*(c)},
\]
the share of the attention of a question that the kept tokens hold, on average over questions.
We estimate it with the questions of each video and average it over the KV heads of all layers and over the videos.
It is below one even for the best choice of $B$ tokens: the best $B$ tokens of each question hold $0.80$, $0.90$ and $0.94$ of its attention on the three backbones.
We report it for two kept sets of each query: \textbf{Single}, one selection on the uncompressed cache, and \textbf{Stream}, the kept set after the last round of the streaming run, where each round scores the cache that earlier rounds left.
The queries are those of Fig.~\ref{fig:v2_gap}; 
\textsc{describe}, recent visual tokens, the chat template and Expected Attention are computed on the stream at every round (computing \textsc{describe} once on the calibration videos instead changes its attention retention by at most $0.010$ after streaming).
Two oracles that know the questions serve as references: \emph{Question known}, where each question compresses the stream on its own and probes with its text at every round, and \emph{Video's questions known}, which probes with all the questions of the video. $\vq_{\mathcal V}$ is a query fitted to the target attention $\va^*$ of each test video.

\begin{table*}[t]
\centering
\small
\setlength{\tabcolsep}{2.8pt}
\begin{tabular}{lcccccccccccc}
\toprule
 & \multicolumn{4}{c}{Qwen2.5-VL} & \multicolumn{4}{c}{Qwen3-VL} & \multicolumn{4}{c}{LLaVA-OV} \\
\cmidrule(lr){2-5}\cmidrule(lr){6-9}\cmidrule(lr){10-13}
Query & KL & Single & Stream & Acc & KL & Single & Stream & Acc & KL & Single & Stream & Acc \\
\midrule
Recent visual tokens & 1.82 & 0.630 & 0.559 & 49.4 & 1.81 & 0.724 & 0.690 & 49.2 & 1.72 & 0.851 & 0.754 & 43.6 \\
Chat template & 1.28 & 0.674 & 0.638 & 49.0 & 1.01 & 0.802 & 0.756 & 53.5 & 1.17 & 0.869 & 0.770 & 42.6 \\
\textsc{describe} & 0.71 & 0.703 & 0.655 & 51.1 & 0.57 & 0.826 & 0.780 & 53.3 & 0.91 & 0.885 & 0.783 & 42.2 \\
Expected Attention & 1.21 & 0.637 & 0.538 & 50.3 & 1.55 & 0.728 & 0.625 & 52.6 & 1.22 & 0.853 & 0.749 & 42.5 \\
TriAttention & 0.62 & 0.701 & 0.560 & 53.2 & 0.57 & 0.828 & 0.700 & 54.2 & 0.79 & 0.896 & 0.788 & 45.2 \\
\midrule
\Ours{} & \textbf{0.41} & \textbf{0.730} & \textbf{0.659} & \textbf{55.9} & \textbf{0.35} & \textbf{0.847} & \textbf{0.786} & \textbf{58.1} & \textbf{0.34} & \textbf{0.914} & \textbf{0.800} & \textbf{45.9} \\
\midrule
\textit{Question known} & -- & 0.796 & 0.715 & 56.5 & -- & 0.897 & 0.821 & 59.0 & -- & 0.940 & 0.818 & 47.5 \\
\textit{Video's questions known} & 0 & 0.786 & 0.713 & 56.4 & 0 & 0.892 & 0.819 & 59.9 & 0 & 0.935 & 0.817 & 47.2 \\
$\vq_{\mathcal V}$ fitted per test video & 0.07 & 0.778 & 0.700 & 56.7 & 0.05 & 0.887 & 0.816 & 59.6 & 0.08 & 0.933 & 0.816 & 46.0 \\
\midrule
Spearman with Acc & $+0.94$ & $+0.77$ & $+0.49$ & -- & $+0.94$ & $+0.94$ & $+0.71$ & -- & $+0.49$ & $+0.49$ & $+0.60$ & -- \\
\bottomrule
\end{tabular}
\caption{\textbf{KL, expected attention retention and accuracy of each query} on MLVU (349 videos, 502 questions). KL: to $\va^*$, on the uncompressed cache. Single, Stream: expected attention retention of one selection on the uncompressed cache and after the last streaming round. Acc: streaming compression at the $6$K budget. Rows below \Ours{} use the questions. Last row: Spearman correlation with Acc over the six queries ($-$KL for KL). Bold: best of the six queries.}
\label{tab:v2_query_full}
\end{table*}

\paragraph{\Ours{} is the closest to the target attention and the most accurate query.}
Across the six queries, a lower KL goes with a higher accuracy, strongly on the Qwen models and more weakly on LLaVA-OV (Table~\ref{tab:v2_query_full}), and \Ours{} has the lowest KL and the highest accuracy on all three backbones.
It is the only one of these queries that is fitted directly to the attention of real questions (Sec.~\ref{sec:v2_analysis}).

\paragraph{Single and Stream.}
Stream is below Single for every query and for both oracles, by $0.03$ to $0.14$ for the queries and by $0.07$ to $0.12$ for the oracles.
The difference is not specific to the surrogate query: the oracles that know the questions show the same order of magnitude, and the difference of \Ours{} ($0.06$ to $0.11$) is within the range of the other queries.
The order of the queries is largely kept: the rank correlation between Single and Stream over the six queries is $0.89$, $0.77$ and $0.94$.
\Ours{} holds the most attention of the six queries on all three backbones, both in a single selection and after streaming.
After streaming, \Ours{} holds $0.92$, $0.96$ and $0.98$ of the attention held by \textit{Question known}.
On the Qwen models, the attention retention after streaming tracks the accuracy less closely than the KL does (Table~\ref{tab:v2_query_full}, last row).

\paragraph{The drop comes from scores that change between rounds.}
If the score of every token were fixed, keeping at each round the top $B$ of the kept tokens and the new chunk would give exactly the top $B$ of the whole stream, since a token kept at the end of any round is among the top $B$ of the prefix by that score; the irreversible eviction would then have no effect.
A constant query vector has this property when the score of a token depends only on the token.
We verify it on the scores of the uncompressed cache: a selection simulated chunk by chunk with these scores holds exactly the attention of the Single selection, for every query.
The difference between Single and Stream therefore comes from the scores changing between rounds.
Three changes of the scores are possible.
\begin{itemize}
\item \textbf{RoPE drift.} The query is rotated to the position at which a question would be appended, which moves with every chunk, so its distance to each cached token differs between rounds and from that of a question at the end of the stream.
\item \textbf{Key drift.} The keys of a new chunk are computed on a cache that earlier rounds have already compressed, so they differ from the keys on the uncompressed cache on which the target attention is defined.
\item \textbf{Normalization.} The score of a token is the average of $S$ softmax maps, each normalized over the tokens currently in the cache, and a KV head takes the maximum of these scores over the query heads that share it (Supp.~\ref{app:gqa}).
The normalizers change as the cache is thinned and differ between the query heads, so the ranking of the tokens can change between rounds even for a fixed query; only for one vector in a single query head is the softmax monotone in the logit and the ranking unchanged.
\end{itemize}
We do not separate their contributions here.
The same changes act on the oracles, which also hold less attention after streaming than in a single selection, so they are properties of the streaming procedure rather than of a particular surrogate query.

\subsection{Why a Constant Query Vector Suffices}
\label{sec:v2_sensitivity}

This subsection extends the analysis of Sec.~\ref{sec:v2_analysis}.
\Ours is a constant vector fixed before deployment, and we ask whether it holds up as the questions, the videos and the stream vary, in KL to $\va^*$ and in MLVU accuracy.

\paragraph{One query for all questions of a video is nearly as accurate as the real question.}
\Ours targets $\va^*$, the average attention of the questions of a video,
since the questions are unknown when the stream is compressed; it
approximates $\va^*$ by fitting to the $\ahatV$ of the calibration videos.
Selecting by $\va^*$, an oracle that uses each video's own questions, is within $0.9$ points of compressing with the real question (Fig.~\ref{fig:v2_gap}), although every question of a video is then answered from the same tokens.
The questions of a video attend alike: since $\va^*$ is their average, the KL of any query from the real questions is their spread around $\va^*$ plus the query's own KL from $\va^*$, and on the 70 MLVU videos with more than one question this spread is only about a quarter of the distance of \Ours{} from the real questions.
The rest is the distance of \Ours{} from $\va^*$, which a better constant query could still reduce.

\paragraph{A per-video query matches the attention far better but improves the accuracy little.}
A query fitted to each video's own $\va^*$, $\vq_{\mathcal V}$ with $S$ vectors per head and the objective of Eq.~\ref{eq:v2_loss}, is an oracle that needs the video's questions, so no deployable method can compute it.
It lowers the KL to $\va^*$ from $0.34$--$0.41$ (\Ours{}) to $0.05$--$0.08$, but raises the accuracy by only $0.8$, $1.5$ and $0.2$ points on the 349 MLVU videos (Fig.~\ref{fig:v2_gap}).
Fitting each video separately thus adds little accuracy beyond one query shared across videos.

\begin{figure}[t]
\centering
\begin{subfigure}[t]{0.49\columnwidth}
    \centering
    \includegraphics[width=\linewidth]{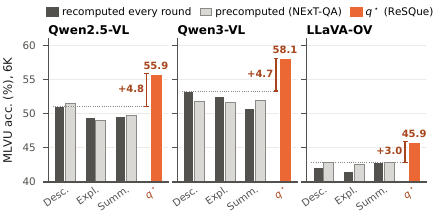}
    \caption{\textbf{Recomputing the prompt query.} MLVU accuracy at 6K with the query of a text prompt recomputed at every round (dark) or precomputed once on NExT-QA (light). $\Delta$: gain of \Ours, also a constant, over the best recomputed prompt.}
    \label{fig:v2_reuse}
\end{subfigure}
\hfill
\begin{subfigure}[t]{0.49\columnwidth}
    \centering
    \includegraphics[width=\linewidth]{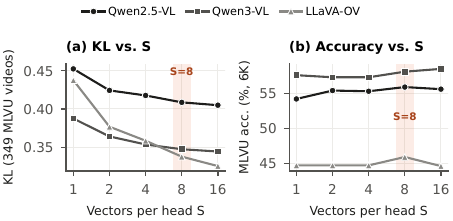}
    \caption{\textbf{Vectors per head.} Effect of the number of query vectors per head $S$ on the attention divergence (left, KL on the 349 MLVU videos) and the MLVU accuracy at the 6K budget (right); \Ours{} uses $S{=}8$ (shaded).}
    \label{fig:v2_analysis}
\end{subfigure}

\caption{\textbf{Additional analysis.}
(a) Recomputing the prompt query.
(b) Effect of the number of query vectors per head.}
\label{fig:merged_analysis}
\end{figure}

\paragraph{Recomputing the prompt query at every round brings no consistent gain.}
Prompt-based compressors recompute the query of their text prompt from the current cache at every round, so that it can adapt to the video seen so far.
Replacing it with a constant, the query of the same prompt computed in advance on the NExT-QA calibration videos, averaged and used at every round without a forward pass, is neither better nor worse (Fig.~\ref{fig:v2_reuse}): across the \textsc{describe}, \textsc{explain} and \textsc{summarize} prompts (Supp.~\ref{app:v2_prompts}) and three backbones, the precomputed query is more accurate than the recomputed one in some settings and less accurate in others, with differences from $-1.6$ to $+1.2$ points and no consistent direction.
Which query is used matters far more: \Ours, also a constant, is $3.0$ to $4.8$ points more accurate than the best of the three recomputed prompts (Fig.~\ref{fig:v2_reuse}, $\Delta$).
This concerns the three generic prompts evaluated here; it does not show that learned probes, such as the soft tokens of Judge Q~\citep{liu2026judgeq} and LookaheadKV~\citep{ahn2026lookaheadkv}, need no recomputation.

\subsection{Alternative Parameterizations}
\label{app:v2_param_ablation}

This subsection extends the comparison of parameterizations of Sec.~\ref{sec:parameterization}.
Fig.~\ref{fig:v2_param_main} compares \Ours{} with one configuration of a neural scorer and one configuration of soft tokens.
Here we vary the main design choices of each alternative to test whether the comparison depends on them.
Every variant is fitted with the same target and objective as \Ours{} (Eq.~\ref{eq:v2_loss}) on the NExT-QA calibration videos, with the training settings of Supp.~\ref{app:v2_fit} and~\ref{app:v2_param_cfg}, and is evaluated on Qwen2.5-VL.
We report the forward KL to $\va^*$ on MLVU (as in Fig.~\ref{fig:v2_gap}), the MLVU accuracy at the 6K budget, and the latency of scoring and selection in one round, measured as in Fig.~\ref{fig:v2_efficiency}.
Both \Ours{} and soft token are timed with their fused kernel implementation (Supp.~\ref{app:v2_inference}, \textit{``Efficient inference with a fused kernel.''}) for fair comparison.

\paragraph{A larger or context-aware scorer does not close the gap to \Ours{}.}
A scorer reads each cached visual token and outputs one score per query head of the group (Supp.~\ref{app:v2_param_cfg}).
We vary its architecture, an MLP that sees each token alone or a two-layer transformer in which a token also attends to the mean key of each chunk of the cache, and its input, the key alone or the key and the value as in KVP~\citep{moschella2026kvp}; all variants are fitted like \Ours{} with Muon for $2{,}000$ steps (Table~\ref{tab:v2_scorer}).
Neither the context of the cache nor the value helps: the transformer, with four times the parameters of the MLP, and the added value both raise the KL and lower the accuracy.
Every scorer has a higher KL and a lower accuracy than \Ours{}, which has fewer parameters than any of them and no per-token network to evaluate.

\paragraph{Scorers read the cache, since keeping hidden states would multiply its memory.}
Fast KVzip~\citep{kim2026fastkvzip} reads the hidden states of each layer.
Re-scoring the cache at every round with such a predictor requires keeping the hidden states of every cached token, which takes $2$ to $3.5$ times the memory of the keys and values.
To compare the parameterizations under a common cache-only setting, we feed our scorers the cached keys alone or together with the values.

\paragraph{More soft-token embeddings stay less accurate than \Ours{}.}
Soft tokens reach the query of each head only through the frozen model, so every round runs a forward pass of the $n$ embeddings, which attends to the whole cache in every layer.
We vary $n\in\{1,2,4,8,16,32\}$, each initialized from the mean embeddings of the NExT-QA~\citep{xiao2021nextqa} questions plus noise and fitted with Muon for $2{,}000$ steps (Table~\ref{tab:v2_softtoken}).
The MLVU KL falls as $n$ grows and is below that of \Ours{} from $n=8$ on, yet the accuracy stays below \Ours{} for every $n$, by $0.5$ points at best ($n=32$).
Even where the soft token fits the attention better, \Ours{} is more accurate for every $n$ and faster: a round takes $93$\,ms against $114$ to $133$\,ms ($1.2$ to $1.4\times$), since every $n$ keeps the forward pass of the embeddings through the model.

\begin{table*}[t]
\centering
\small
\setlength{\tabcolsep}{4pt}
\begin{tabular}{lccccc}
\toprule
$n$ & Params & Train KL & MLVU KL & Acc & ms / round \\
\midrule
1  & 3.6K  & 0.325 & 0.685 & 54.3 & 114 \\
2  & 7.2K  & 0.215 & 0.542 & 54.7 & 116 \\
4  & 14.3K & 0.132 & 0.451 & 54.5 & 115 \\
8  & 28.7K & 0.072 & 0.369 & 54.3 & 119 \\
16 & 57.3K & 0.051 & 0.313 & 54.9 & 124 \\
\textbf{32 (default)} & 115K  & 0.063 & 0.315 & 55.4 & 133 \\
\midrule
\textbf{\Ours{} ($S{=}8$)} & 0.80M & 0.052 & 0.409 & 55.9 & 93 \\
\bottomrule
\end{tabular}
\caption{\textbf{Soft tokens with $n$ embeddings} on Qwen2.5-VL. Train KL is measured on the NExT-QA calibration videos, MLVU KL and accuracy on MLVU at 6K; ms / round: round latency, measured as in Fig.~\ref{fig:v2_efficiency}.}
\label{tab:v2_softtoken}
\end{table*}

\begin{table*}[t]
\centering
\small
\begin{tabular}{llccccc}
\toprule
Scorer & Input & Params & Train KL & MLVU KL & Acc & ms / round \\
\midrule
\textbf{MLP (default)} & K    & 0.85M & 0.067 & 0.414 & 54.7 & 301 \\
MLP         & K, V & 1.31M & 0.069 & 0.418 & 54.3 & 329 \\
Transformer & K    & 3.5M  & 0.088 & 0.554 & 54.3 & 298 \\
Transformer & K, V & 4.0M  & 0.129 & 0.660 & 53.6 & 299 \\
\midrule
\textbf{\Ours{} }    & K    & 0.80M & 0.052 & 0.409 & 55.9 & 93 \\
\bottomrule
\end{tabular}
\caption{\textbf{Neural scorers} on Qwen2.5-VL, fitted with the same target and objective as \Ours{} and with Muon; each scorer row is the mean of two fits with different seeds. Train KL is measured on the NExT-QA calibration videos and MLVU KL on the held-out MLVU videos; ms / round: round latency, measured as in Fig.~\ref{fig:v2_efficiency}; the differences among the scorers are within the measurement variation.}
\label{tab:v2_scorer}
\end{table*}

\section{Ablation Study of \Ours}
\label{sec:v2_head}

This section ablates the design choices of \Ours{} in the query, the objective, the starting point, the calibration data and the aggregation under grouped-query attention, extending the analysis of Sec.~\ref{sec:v2_analysis}.

\subsection{Query}
\label{app:v2_abl_query}

\begin{table}[t]
\centering
\small
\setlength{\tabcolsep}{4pt}
\begin{tabular}{@{}llrrr@{}}
\toprule
Query & Queries & Qwen2.5-VL & Qwen3-VL & LLaVA-OV \\
\midrule
\Ours{} (per layer and head) & $LH$ & 54.2 & 57.6 & 44.7 \\
\midrule
\quad shared within a KV group & $LH_{\mathrm{kv}}$ & $-1.1$ & $-0.6$ & $-0.1$ \\
\quad shared within a layer & $L$ & $-7.1$ & $-8.7$ & $-0.1$ \\
\quad shared across layers & $H$ & $-8.1$ & $-9.3$ & $-2.4$ \\
\quad shared by the whole model & $1$ & $-9.4$ & $-13.1$ & $-2.4$ \\
\bottomrule
\end{tabular}
\caption{\textbf{Sharing the query across heads.} MLVU accuracy (\%, 6K) of \Ours{} with one vector per head ($S{=}1$), and the change when one query is shared by the heads of a KV group, of a layer, with the same head index across layers, or by the whole model; each variant is fitted with the same objective and data. Queries: number of distinct queries ($L$ layers, $H$ query heads and $H_{\mathrm{kv}}$ KV heads per layer).}
\label{tab:v2_sharing_drop}
\end{table}

Within the query, we ablate how many vectors each head holds and how widely a query is shared across heads.
All variants are fitted with the objective of Eq.~\ref{eq:v2_loss} on the same 570 NExT-QA calibration videos.

\begin{figure*}[t]
\centering
\includegraphics[width=\textwidth]{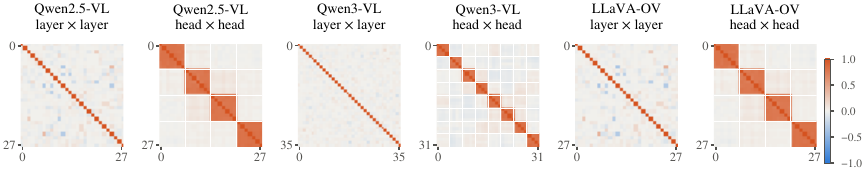}
\caption{\textbf{Similarity of the fitted queries across layers and heads.} Cosine similarity between fitted queries (\Ours with $S{=}1$). Layer $\times$ layer: between two layers, each represented by the queries of all its heads concatenated. Head $\times$ head: between two heads of the same layer, averaged over layers; white lines separate KV groups.}
\label{fig:v2_heads}
\end{figure*}

\paragraph{A few vectors per head suffice: the KL saturates by $S=8$.}
The query of a head holds $S$ vectors whose maps are averaged (Eq.~\ref{eq:v2_rows}); Fig.~\ref{fig:v2_analysis} varies $S$ from $1$ to $16$.
The KL decreases with every added vector on all three backbones and is nearly saturated by $S=8$.
More vectors expand the representational capacity of the surrogate, and they do not collapse into one: after the fit, the $S=8$ vectors of a head diverge (mean pairwise cosine about $0.8$, against about $0.99$ at the start).
$S=8$ improves the accuracy over $S=1$ on every backbone, while $S=16$ is not consistently better; we chose $S=8$ because the KL has almost stopped decreasing by then, without claiming that it is the best value for every backbone.

\begin{table}[t]
\centering
\small
\setlength{\tabcolsep}{5pt}
\begin{tabular}{lccccc}
\toprule
Query & Vectors & Fit & Qwen2.5-VL & Qwen3-VL & LLaVA-OV \\
\midrule
Mean question query $\qbar$ & 1 & no & 53.3 & 55.6 & 44.5 \\
$k$-means centroids & 8 & no & 53.9 & 55.8 & 44.4 \\
Fitted, $S{=}1$ & 1 & yes & 54.2 & 57.6 & 44.7 \\
\textbf{Fitted, $S{=}8$ (\Ours{})} & 8 & yes & \textbf{55.9} & \textbf{58.1} & \textbf{45.9} \\
\bottomrule
\end{tabular}
\caption{\textbf{Fitting versus the number of vectors.} MLVU accuracy (\%, 6K) of a query without and with the fit and with one or eight vectors per head. $\qbar$ is the mean of the question queries of the NExT-QA pool; the centroids are the $8$ $k$-means centroids of the token queries of NExT-QA questions, computed for each head; the fitted rows are the $S{=}1$ and $S{=}8$ points of Fig.~\ref{fig:v2_analysis}.}
\label{tab:v2_kmeans}
\end{table}

\paragraph{Extra vectors help only when they are fitted.}
Table~\ref{tab:v2_kmeans} asks whether the gain of $S=8$ over $S=1$ comes from the additional vectors or from fitting them.
Without a fit, the $8$ $k$-means centroids are within $0.6$ points of the mean question query $\qbar$ on all three backbones and $0.3$, $1.8$ and $0.3$ points below the fit with $S=1$, whereas the fit with $S=8$ is $2.0$, $2.3$ and $1.5$ points above the centroids.
Vectors added without a fit therefore do not help; they pay off only when they are fitted to the target attention, where $S=8$ is $1.7$, $0.5$ and $1.2$ points above $S=1$.

\paragraph{Queries must be head-specific; sharing beyond a KV group lowers the accuracy.}
With one vector per head ($S=1$), where sharing across heads is well defined, we fit queries shared by a KV group, a layer, a head index across layers, or the whole model with the same objective and data (Table~\ref{tab:v2_sharing_drop}).
Sharing beyond a KV group lowers the MLVU accuracy by $7$ to $13$ points on the Qwen models and by up to $2.4$ points on LLaVA-OV, whereas sharing within a KV group changes it by about one point or less.
The fitted queries are similar within a KV group but nearly orthogonal across groups on all three backbones (Fig.~\ref{fig:v2_heads}), consistent with the small effect of sharing within a group.

\subsection{Objective of the Fit}
\label{app:v2_divergence}
\begin{figure*}[t]
\centering
\begin{subfigure}[t]{0.48\textwidth}
    \centering
    \includegraphics[width=\linewidth]{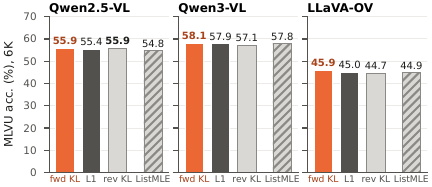}
    \caption{\textbf{Objective of attention distribution matching.}
    MLVU accuracy at the 6K budget with $\operatorname{Div}$ in Eq.~\ref{eq:v2_loss} the forward KL divergence, the $L_1$ distance, or the reverse KL divergence, or with the ranking loss ListMLE (hatched); $S{=}8$ vectors per head, fitted with Muon~\citep{jordan2024muon} on the 570 NExT-QA calibration videos, one fit seed. Bold: best per model.}
    \label{fig:v2_div}
\end{subfigure}
\hfill
\begin{subfigure}[t]{0.48\textwidth}
    \centering
    \includegraphics[width=\linewidth]{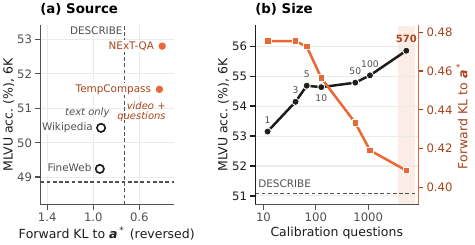}
    \caption{\textbf{Calibration data.}
    Left, source: \Ours fitted on 100 samples of each source, a video with two questions or, for a text source, a passage with the text that follows it, averaged over the backbones. Closer to the questions is to the right.
    Right, size: fits on the first $N$ NExT-QA videos on Qwen2.5-VL ($N$ next to each point, with $N{=}570$ the \Ours of the main experiments). Dashed: \textsc{describe}.}
    \label{fig:v2_calib}
\end{subfigure}

\caption{\textbf{Additional ablations.}
(a) Effect of the objective for attention distribution matching: divergences and a ranking loss.
(b) Effect of calibration data source and size.}
\label{fig:v2_ablations}
\end{figure*}

\paragraph{The choice of divergence matters little.}
The objective of Eq.~\ref{eq:v2_loss} admits any divergence $\operatorname{Div}$; we use the forward KL.
Fig.~\ref{fig:v2_div} compares it with the $L_1$ distance and the reverse KL $\mathrm{KL}(\va_\theta(\mathcal V)\,\|\,\ahatV)$, and with the ranking loss ListMLE~\citep{xia2008listwise}, which fits the order of the top $6$K tokens of $\ahatV$, as many as the $6$K budget keeps, rather than the distribution, with $\log\va_\theta(\mathcal V)$ as the scores; each objective is fitted in the setting of \Ours ($S=8$ vectors per head).
The three divergences give similar accuracy, since each is minimized when the predicted attention equals the target, whose ranking gives the optimal selection of Sec.~\ref{sec:v2_target}.
ListMLE, which constrains only the order of the top tokens, is slightly less accurate than the forward KL on every backbone, and we adopt the forward KL as our default objective.

\subsection{Initialization of the Fit}
\label{app:v2_init}

This subsection compares the starting points of the fit of Supp.~\ref{app:v2_fit}.

\paragraph{The fit, not the starting point, determines the accuracy.}
We compare three starts of the fit of \Ours{} ($S{=}8$): the mean question query $\qbar$ (Supp.~\ref{app:v2_fit}), the query of \textsc{describe}, and the constant vector with every element $1/\sqrt{d}$, which carries no information about the model or the video; each start is perturbed and optimized as in Supp.~\ref{app:v2_fit}, and the rows of $\qbar$ after the fit are \Ours{} of Table~\ref{tab:v2_main} (Table~\ref{tab:v2_init_starts}).
Before the fit, the constant vector is far from $\va^*$ and up to $11$ points less accurate than the other two starts; after the fit, the three starts are within $1$ point of each other on every backbone.
The objective pulls every start toward the same target attention, so the starting point matters little for the accuracy.
It still matters for the KL on Qwen3-VL and LLaVA-OV, where the constant start ends farther from $\va^*$, so we keep $\qbar$ as the default.

\begin{table}[t]
\centering
\footnotesize
\setlength{\tabcolsep}{4pt}
\begin{tabular}{@{}l l cc cc@{}}
\toprule
 & & \multicolumn{2}{c}{before the fit} & \multicolumn{2}{c}{after the fit}\\
\cmidrule(lr){3-4}\cmidrule(lr){5-6}
Backbone & Start & KL $\downarrow$ & Acc $\uparrow$ & KL $\downarrow$ & Acc $\uparrow$\\
\midrule
\multirow{3}{*}{Qwen2.5-VL} & $\qbar$ & 0.81 & 53.3 & 0.41 & 55.9\\
 & \textsc{describe} & 0.68 & 51.5 & 0.42 & 55.2\\
 & $1/\sqrt{d}$ & 1.75 & 44.8 & 0.42 & 55.7\\
\midrule
\multirow{3}{*}{Qwen3-VL} & $\qbar$ & 0.56 & 55.6 & 0.35 & 58.1\\
 & \textsc{describe} & 0.62 & 51.8 & 0.35 & 57.6\\
 & $1/\sqrt{d}$ & 3.72 & 44.3 & 0.49 & 57.3\\
\midrule
\multirow{3}{*}{LLaVA-OV} & $\qbar$ & 0.83 & 44.5 & 0.34 & 45.9\\
 & \textsc{describe} & 0.83 & 42.8 & 0.40 & 45.3\\
 & $1/\sqrt{d}$ & 3.02 & 41.7 & 0.55 & 45.0\\
\bottomrule
\end{tabular}
\caption{\textbf{Starting point of the fit.} \Ours{} ($S{=}8$) fitted from three starts. KL to $\va^*$ (349 MLVU videos) and MLVU accuracy (\%, Acc) at the 6K budget, of the start used as the query (before the fit) and of the fitted query (after the fit).}
\label{tab:v2_init_starts}
\end{table}

\subsection{Calibration Data}
\label{app:v2_calib}

The query parameters are fitted exclusively on calibration data disjoint from the evaluation benchmarks(Supp.~\ref{app:v2_abl_query}). We vary the source of the calibration data (domain and task) and then its amount.

\paragraph{Calibration needs videos with questions; generic text helps little.}
The calibration set of \Ours pairs videos with questions, as the model meets them at inference.
We fit \Ours with the same procedure on 100 samples of each of four sources: two VideoQA sets, NExT-QA and TempCompass~\citep{liu2024tempcompass} (questions on direction, speed and order), and two public corpora of general text, FineWeb~\citep{penedo2024fineweb} and Wikipedia (Fig.~\ref{fig:v2_calib}, left).
Averaged over the three backbones, both video sources are closer to $\va^*$ than \textsc{describe} (KL about $0.4$ against $0.73$) and more accurate, whereas the text corpora are farther from it than \textsc{describe} and only slightly more accurate.
Text calibration differs from deployment in two respects: it places a passage, not a video, in the context, and it fits the continuation of the passage rather than a question. Among the video sources, TempCompass is not consistently more accurate than NExT-QA.

\paragraph{Ten videos already recover most of the gain.}
We fit \Ours with Muon for $2{,}000$ steps on the first $N$ NExT-QA videos with all their questions, on Qwen2.5-VL (Fig.~\ref{fig:v2_calib}, right).
Ten videos already close $84\%$ of the KL distance from \textsc{describe} to the fit on all 570 videos; from $N=10$ to $N=100$ the accuracy stays nearly flat while the KL keeps falling slowly, and all 570 videos add $0.8$ points.
A small calibration set thus suffices to fit the query, and more videos refine it gradually.

\subsection{Aggregation under Grouped-Query Attention}
\label{app:gqa}
\paragraph{Max over query heads is slightly better than mean.}
The query heads that share a KV head each give a score to every cached token, and the KV head keeps the $B$ tokens with the largest maximum of these scores (Supp.~\ref{app:v2_inference}).
Replacing the maximum by the mean over the query heads, with the same query and protocol (Table~\ref{tab:v2_gqa}), lowers the accuracy in $21$ of the $27$ settings of Table~\ref{tab:v2_main}, but by at most $1.1$ points averaged over the three benchmarks.
The maximum is an empirical design choice: the expected attention retention of Sec.~\ref{sec:v2_target}, summed over the query heads of a group, would instead call for their sum or mean.
We use maximum aggregation as a fixed default across all backbones and budgets.
\begin{table}[t]
\centering
\footnotesize
\setlength{\tabcolsep}{2.5pt}
\resizebox{0.60\columnwidth}{!}{%
\begin{tabular}{ll cc cc cc}
\toprule
 & & \multicolumn{2}{c}{6K} & \multicolumn{2}{c}{3K} & \multicolumn{2}{c}{1K}\\
\cmidrule(lr){3-4}\cmidrule(lr){5-6}\cmidrule(lr){7-8}
Backbone & Benchmark & max & mean & max & mean & max & mean\\
\midrule
\multirow{4}{*}{Qwen2.5-VL} & MLVU & \textbf{55.9} & 55.3 & \textbf{52.9} & 52.3 & \textbf{50.2} & 48.2\\
 & Video-MME & \textbf{59.1} & 57.9 & \textbf{57.2} & 56.9 & \textbf{54.4} & 53.8\\
 & LongVideoBench & \textbf{62.4} & 61.7 & 60.7 & \textbf{60.8} & \textbf{59.4} & 58.7\\
 & Overall & \textbf{59.1} & 58.3 & \textbf{56.9} & 56.7 & \textbf{54.7} & 53.6\\
\midrule
\multirow{4}{*}{Qwen3-VL} & MLVU & \textbf{58.1} & 56.9 & \textbf{55.1} & 54.3 & \textbf{47.8} & 47.3\\
 & Video-MME & \textbf{62.9} & 62.4 & 60.2 & \textbf{60.5} & \textbf{57.9} & 57.7\\
 & LongVideoBench & \textbf{62.6} & 62.1 & \textbf{61.1} & 60.5 & \textbf{58.6} & 58.2\\
 & Overall & \textbf{61.2} & 60.5 & \textbf{58.8} & 58.4 & \textbf{54.7} & 54.4\\
\midrule
\multirow{4}{*}{LLaVA-OV} & MLVU & \textbf{45.9} & 45.1 & 44.0 & \textbf{44.8} & \textbf{41.8} & 41.6\\
 & Video-MME & \textbf{50.2} & \textbf{50.2} & \textbf{48.9} & \textbf{48.9} & \textbf{47.9} & 47.5\\
 & LongVideoBench & \textbf{57.8} & \textbf{57.8} & \textbf{57.9} & 57.2 & \textbf{57.2} & 55.5\\
 & Overall & \textbf{51.3} & 51.0 & \textbf{50.3} & \textbf{50.3} & \textbf{49.0} & 48.2\\
\bottomrule
\end{tabular}}
\caption{\textbf{Aggregating the query heads of a KV group.} Accuracy (\%) of \Ours{} when a KV head scores each token by the maximum (default) or the mean of the scores of the query heads that share it, at about 6K / 3K / 1K visual tokens per head; the query is the same. Overall: mean over the three benchmarks. Bold: the higher of the two.}
\label{tab:v2_gqa}
\end{table}

\section{Limitations and Future Work}
\label{app:limitations}

This section states the limitations of \Ours{} and the directions we leave for future work.

\paragraph{Limitations.}
First, \Ours{} prunes tokens and does nothing else to the cache.
Methods that merge evicted tokens into the kept ones~\citep{zhang2024cam,wan2024d2o} or quantize the cache~\citep{yuan2024kivi,hooper2024kvquant} are orthogonal to it, and combining them with \Ours{} may give better accuracy at the same memory; we have not tested such a combination.
Second, \Ours{} scores a token by its attention weight alone, which does not reflect what the token contributes to the output.
The contribution of a token to the output of a head is its attention weight multiplied by its value vector, so two tokens with the same weight are ranked alike whatever their values.
Value-aware scores~\citep{guo2024vatp} include the value by weighting the attention with the norm of the value vector.

\paragraph{Future work.}
We outline two directions.
First, \Ours{} gives every head the same budget, so that the queries we compare differ only in the tokens they select.
Methods that allocate the budget across heads, by the concentration of their attention~\citep{feng2024adakv} or to heads identified offline as retrieval or visual heads~\citep{tang2025duoattention,wang2025sparsemm,park2026vrh}, are orthogonal to the choice of query and could be applied on top of \Ours{}.
Second, a value term such as the value norm of VATP~\citep{guo2024vatp} can be added to the score of \Ours{}; we have not tested whether it improves the selection.

\onecolumn
\section{Full Results}
\label{app:full_results}
\renewcommand{\dbltopfraction}{0.95}\renewcommand{\textfraction}{0.05}\renewcommand{\floatpagefraction}{0.8}

This section gives the numbers behind the figures of the main paper: the accuracy of every compressor on the long-video benchmarks (Supp.~\ref{app:v2_main_results}), its breakdown by task group, video duration and video length and the streaming benchmark (Supp.~\ref{app:v2_per_category}), and the accuracy of every query of the query analysis at the three budgets (Supp.~\ref{app:v2_query_budgets}).

\begin{table*}[!t]
\centering
\normalsize
\setlength{\tabcolsep}{6pt}
\renewcommand{\arraystretch}{0.95}
\begin{tabular}{l cccc}
\toprule
\textbf{Method} & \textbf{MLVU} & \textbf{Video-MME} & \textbf{LongVideoBench} & \textbf{Overall}\\
\midrule
\multicolumn{5}{l}{\emph{Qwen2.5-VL-7B}}\\
\midrule
Full KV & 54.0 & 59.4 & 63.9 & 59.1\\
\midrule
SWA & 45.9/39.9/40.8 & 50.9/48.8/47.0 & 60.0/57.6/57.3 & 52.3/48.8/48.4\\
InfiniPot-V & 50.3/47.7/43.4 & 56.9/54.6/52.3 & 62.0/60.3/58.2 & 56.4/54.2/51.3\\
CoRDS & 48.6/47.4/42.4 & 54.6/52.8/49.4 & 61.9/59.2/57.9 & 55.0/53.1/49.9\\
KeyDiff & 52.3/48.6/42.4 & 55.3/52.8/50.8 & 62.0/\textbf{60.7}/58.4 & 56.6/54.0/50.5\\
StreamMem & 49.7/45.2/44.3 & 56.5/54.2/49.8 & 60.3/58.0/58.3 & 55.5/52.5/50.8\\
HERMES & 53.0/49.7/42.9 & 57.2/55.1/51.8 & 60.6/59.5/58.7 & 56.9/54.7/51.1\\
Expected Attention & 48.4/48.9/44.9 & 56.6/54.5/51.6 & 61.9/59.5/58.0 & 55.6/54.3/51.5\\
TriAttention & \underline{53.2}/\underline{50.8}/\underline{49.4} & \underline{58.2}/\underline{55.8}/\underline{53.7} & \underline{62.3}/\underline{60.4}/\underline{59.3} & \underline{57.9}/\underline{55.6}/\underline{54.2}\\
\midrule
\Ours{} & \textbf{55.9}/\textbf{52.9}/\textbf{50.2} & \textbf{59.1}/\textbf{57.2}/\textbf{54.4} & \textbf{62.4}/\textbf{60.7}/\textbf{59.4} & \textbf{59.1}/\textbf{56.9}/\textbf{54.7}\\
\midrule
\multicolumn{5}{l}{\emph{Qwen3-VL-8B}}\\
\midrule
Full KV & 62.4 & 65.2 & 64.5 & 64.0\\
\midrule
SWA & 41.4/39.9/38.1 & 52.4/50.3/47.5 & 56.5/54.8/52.0 & 50.1/48.4/45.9\\
InfiniPot-V & \underline{54.6}/48.0/43.1 & 61.4/\underline{58.7}/53.1 & \underline{60.9}/58.7/54.8 & 58.9/55.1/50.4\\
CoRDS & 51.9/50.1/43.1 & 57.8/54.3/49.6 & 60.5/58.7/53.1 & 56.7/54.4/48.6\\
KeyDiff & 48.2/45.5/39.3 & 54.9/53.2/47.6 & 58.8/55.5/49.6 & 54.0/51.4/45.5\\
StreamMem & 50.0/46.0/39.7 & 59.1/55.0/49.1 & 59.6/57.3/54.3 & 56.3/52.8/47.7\\
HERMES & 48.6/44.1/41.2 & 57.6/55.3/50.1 & 59.1/58.0/\underline{55.5} & 55.1/52.5/48.9\\
Expected Attention & 50.8/46.1/43.2 & 60.7/57.5/\underline{53.9} & \underline{60.9}/58.3/54.9 & 57.5/54.0/50.7\\
TriAttention & 54.2/\underline{51.0}/\underline{46.3} & \underline{62.0}/58.1/53.6 & 60.7/\underline{59.3}/55.1 & \underline{59.0}/\underline{56.1}/\underline{51.7}\\
\midrule
\Ours{} & \textbf{58.1}/\textbf{55.1}/\textbf{47.8} & \textbf{62.9}/\textbf{60.2}/\textbf{57.9} & \textbf{62.6}/\textbf{61.1}/\textbf{58.6} & \textbf{61.2}/\textbf{58.8}/\textbf{54.7}\\
\midrule
\multicolumn{5}{l}{\emph{LLaVA-OV-7B}}\\
\midrule
Full KV & 47.6 & 50.6 & 58.8 & 52.3\\
\midrule
SWA & 40.3/38.6/36.7 & 48.2/46.0/45.0 & 57.3/56.5/56.0 & 48.6/47.0/45.9\\
InfiniPot-V & 42.7/40.3/38.6 & 49.4/47.9/46.7 & \textbf{58.2}/57.2/\textbf{57.2} & 50.1/48.5/47.5\\
CoRDS & 42.8/40.8/37.3 & \underline{49.5}/47.1/45.2 & 57.6/56.7/55.5 & 50.0/48.2/46.0\\
KeyDiff & 44.8/42.6/40.0 & 48.1/47.7/46.1 & \textbf{58.2}/\underline{57.6}/55.7 & 50.4/49.3/47.3\\
StreamMem & 43.1/42.3/38.5 & 48.8/\underline{48.1}/\underline{47.2} & 57.3/56.9/55.8 & 49.7/49.1/47.2\\
HERMES & 41.0/39.3/37.0 & 48.3/47.3/44.9 & 57.1/56.9/\underline{56.2} & 48.8/47.8/46.1\\
Expected Attention & 43.6/41.7/38.5 & 49.1/47.7/46.5 & \underline{57.9}/57.0/54.9 & 50.2/48.8/46.6\\
TriAttention & \underline{45.2}/\textbf{44.8}/\underline{41.6} & 49.4/48.0/46.8 & \underline{57.9}/57.0/55.7 & \underline{50.8}/\underline{49.9}/\underline{48.0}\\
\midrule
\Ours{} & \textbf{45.9}/\underline{44.0}/\textbf{41.8} & \textbf{50.2}/\textbf{48.9}/\textbf{47.9} & 57.8/\textbf{57.9}/\textbf{57.2} & \textbf{51.3}/\textbf{50.3}/\textbf{49.0}\\
\bottomrule
\end{tabular}
\caption{\textbf{Long-video question answering under streaming compression.} Each cell: accuracy (\%) at budgets of about 6K\,/\,3K\,/\,1K visual tokens per head. MLVU: average over task groups; Video-MME: average over medium and long videos; LongVideoBench: average over the four video-length groups of the validation set; Overall: mean of the three benchmarks. \textbf{Bold}: best, \underline{underline}: second best per budget, column and backbone, Full KV excluded. \Ours{}: fitted query $\qstar$ with eight vectors per head.}
\label{tab:v2_main}
\end{table*}

\subsection{Main Results}
\label{app:v2_main_results}

Table~\ref{tab:v2_main} gives the accuracy of every compressor on MLVU, Video-MME and LongVideoBench at the 6K, 3K and 1K budgets on the three backbones, with Full KV as reference: the numbers from which the comparison of Fig.~\ref{fig:v2_main} is drawn.

\clearpage
\begin{table*}[!t]
\centering
\normalsize
\setlength{\tabcolsep}{4pt}
\renewcommand{\arraystretch}{0.95}
\begin{tabular}{l cccc ccc cccc}
\toprule
 & \multicolumn{4}{c}{MLVU} & \multicolumn{3}{c}{Video-MME} & \multicolumn{4}{c}{LongVideoBench}\\
\cmidrule(lr){2-5}\cmidrule(lr){6-8}\cmidrule(lr){9-12}
Method & Single & Multi & Holistic & Avg & Medium & Long & Avg & $\leq$60\,s & 600\,s & 3600\,s & Avg\\
\midrule
\multicolumn{12}{l}{\emph{Qwen2.5-VL-7B}}\\
Full KV & 48.0 & 38.1 & 75.9 & 54.0 & 64.7 & 54.2 & 59.4 & 69.8 & 63.6 & 52.3 & 63.9\\
\midrule
SWA & 36.3 & 35.4 & 66.0 & 45.9 & 52.4 & 49.3 & 50.9 & 70.9 & 50.5 & 47.7 & 60.0\\
InfiniPot-V & 45.6 & 33.6 & 71.6 & 50.3 & 60.9 & 53.0 & 56.9 & 70.6 & \underline{58.5} & 48.0 & 62.0\\
CoRDS & 41.1 & 33.6 & 70.9 & 48.6 & 57.6 & 51.6 & 54.6 & \underline{71.2} & 57.3 & 47.9 & 61.9\\
KeyDiff & 42.3 & 38.1 & \textbf{76.6} & 52.3 & 58.3 & 52.3 & 55.3 & \underline{71.2} & 55.1 & \textbf{50.5} & 62.0\\
StreamMem & 42.3 & 33.6 & 73.0 & 49.7 & 60.0 & 53.0 & 56.5 & \underline{71.2} & 53.2 & 45.7 & 60.3\\
HERMES & 44.0 & \textbf{43.4} & 71.6 & 53.0 & 58.9 & \underline{55.4} & 57.2 & 70.9 & 52.9 & 47.7 & 60.6\\
Expected Attention & 44.0 & 31.9 & 69.5 & 48.4 & 59.8 & 53.3 & 56.6 & \textbf{71.5} & 57.0 & 47.5 & 61.9\\
TriAttention & \underline{47.6} & 38.9 & 73.0 & \underline{53.2} & \textbf{62.6} & 53.8 & \underline{58.2} & \textbf{71.5} & 56.8 & \underline{49.3} & \underline{62.3}\\
\midrule
\Ours{} & \textbf{49.2} & \underline{42.5} & \underline{75.9} & \textbf{55.9} & \underline{61.7} & \textbf{56.4} & \textbf{59.1} & 70.6 & \textbf{59.5} & 48.6 & \textbf{62.4}\\
\midrule
\multicolumn{12}{l}{\emph{Qwen3-VL-8B}}\\
Full KV & 55.6 & 51.3 & 80.1 & 62.4 & 69.2 & 61.1 & 65.2 & 67.6 & 66.0 & 56.9 & 64.5\\
\midrule
SWA & 32.3 & 31.0 & 61.0 & 41.4 & 53.0 & 51.8 & 52.4 & \underline{69.0} & 45.9 & 42.4 & 56.5\\
InfiniPot-V & \underline{50.0} & \textbf{44.2} & 69.5 & \underline{54.6} & \underline{65.7} & 57.1 & 61.4 & 68.7 & 59.0 & \underline{47.3} & \underline{60.9}\\
CoRDS & 47.2 & 39.8 & 68.8 & 51.9 & 61.8 & 53.9 & 57.8 & 67.3 & 60.9 & 46.5 & 60.5\\
KeyDiff & 43.1 & 35.4 & 66.0 & 48.2 & 57.3 & 52.6 & 54.9 & 68.7 & 53.2 & 45.0 & 58.8\\
StreamMem & 45.6 & 33.6 & 70.9 & 50.0 & 62.3 & 55.9 & 59.1 & \underline{69.0} & 55.6 & 45.4 & 59.6\\
HERMES & 39.9 & 39.8 & 66.0 & 48.6 & 58.9 & 56.3 & 57.6 & 67.3 & 55.3 & 46.5 & 59.1\\
Expected Attention & 45.2 & 36.3 & 70.9 & 50.8 & 64.2 & 57.1 & 60.7 & \textbf{69.3} & 58.7 & 46.6 & \underline{60.9}\\
TriAttention & \underline{50.0} & 38.9 & \underline{73.8} & 54.2 & 65.0 & \underline{59.0} & \underline{62.0} & 67.0 & \underline{61.9} & 47.0 & 60.7\\
\midrule
\Ours{} & \textbf{52.8} & \underline{43.4} & \textbf{78.0} & \textbf{58.1} & \textbf{65.9} & \textbf{60.0} & \textbf{62.9} & \textbf{69.3} & \textbf{62.9} & \textbf{49.1} & \textbf{62.6}\\
\midrule
\multicolumn{12}{l}{\emph{LLaVA-OV-7B}}\\
Full KV & 43.1 & 23.0 & 76.6 & 47.6 & 54.7 & 46.6 & 50.6 & 66.2 & 53.4 & 49.3 & 58.8\\
\midrule
SWA & 33.5 & 22.1 & 65.2 & 40.3 & 51.7 & 44.8 & 48.2 & \underline{66.5} & 49.8 & \underline{46.5} & 57.3\\
InfiniPot-V & 35.1 & 22.1 & 70.9 & 42.7 & 51.6 & \textbf{47.3} & 49.4 & \underline{66.5} & \underline{53.4} & 46.3 & \textbf{58.2}\\
CoRDS & 36.3 & \textbf{24.8} & 67.4 & 42.8 & 52.2 & \underline{46.8} & \underline{49.5} & \underline{66.5} & 52.2 & 45.2 & 57.6\\
KeyDiff & 38.3 & \underline{23.9} & 72.3 & 44.8 & 50.7 & 45.4 & 48.1 & \textbf{66.8} & \textbf{54.1} & 45.2 & \textbf{58.2}\\
StreamMem & 37.1 & 21.2 & 70.9 & 43.1 & \underline{52.6} & 45.0 & 48.8 & \underline{66.5} & 51.2 & 45.0 & 57.3\\
HERMES & 33.1 & \textbf{24.8} & 65.2 & 41.0 & 51.2 & 45.3 & 48.3 & \underline{66.5} & 50.5 & 45.0 & 57.1\\
Expected Attention & 35.1 & \textbf{24.8} & 70.9 & 43.6 & 52.3 & 45.8 & 49.1 & \textbf{66.8} & 53.2 & 44.9 & \underline{57.9}\\
TriAttention & \textbf{39.5} & 23.0 & \underline{73.0} & \underline{45.2} & \underline{52.6} & 46.2 & 49.4 & \underline{66.5} & 52.9 & 45.7 & \underline{57.9}\\
\midrule
\Ours{} & \underline{39.1} & \textbf{24.8} & \textbf{73.8} & \textbf{45.9} & \textbf{54.0} & 46.4 & \textbf{50.2} & 65.9 & 52.7 & \textbf{46.6} & 57.8\\
\bottomrule
\end{tabular}
\caption{\textbf{Task-level accuracy at the 6K budget.} The numbers of Table~\ref{tab:v2_main} split by task group (MLVU), video duration (Video-MME medium and long videos) and video length (LongVideoBench: videos of at most 60, 600 and 3600 seconds; the 15 and 60 second groups are merged), in \%; Avg is the value of the main table, the macro average (unweighted mean) of the groups of each benchmark, which for LongVideoBench are the four length groups of 15, 60, 600 and 3600 seconds. Bold: best, underlined: second best per column and backbone.}
\label{tab:v2_main_ext_6k}
\end{table*}

\begin{table*}[!t]
\centering
\normalsize
\setlength{\tabcolsep}{4pt}
\renewcommand{\arraystretch}{0.95}
\begin{tabular}{l cccc ccc cccc}
\toprule
 & \multicolumn{4}{c}{MLVU} & \multicolumn{3}{c}{Video-MME} & \multicolumn{4}{c}{LongVideoBench}\\
\cmidrule(lr){2-5}\cmidrule(lr){6-8}\cmidrule(lr){9-12}
Method & Single & Multi & Holistic & Avg & Medium & Long & Avg & $\leq$60\,s & 600\,s & 3600\,s & Avg\\
\midrule
\multicolumn{12}{l}{\emph{Qwen2.5-VL-7B}}\\
Full KV & 48.0 & 38.1 & 75.9 & 54.0 & 64.7 & 54.2 & 59.4 & 69.8 & 63.6 & 52.3 & 63.9\\
\midrule
SWA & 30.2 & 25.7 & 63.8 & 39.9 & 50.0 & 47.6 & 48.8 & 68.1 & 48.8 & 45.7 & 57.6\\
InfiniPot-V & 40.7 & 33.6 & 68.8 & 47.7 & 58.7 & 50.6 & 54.6 & \underline{70.6} & 53.6 & 46.3 & 60.3\\
CoRDS & 39.5 & 31.9 & 70.9 & 47.4 & 55.8 & 49.8 & 52.8 & 69.5 & 52.2 & 45.6 & 59.2\\
KeyDiff & 37.1 & 32.7 & \textbf{75.9} & 48.6 & 56.1 & 49.4 & 52.8 & 69.8 & 53.9 & \textbf{49.5} & \textbf{60.7}\\
StreamMem & 35.9 & 31.0 & 68.8 & 45.2 & 56.1 & 52.3 & 54.2 & 68.7 & 51.2 & 43.6 & 58.0\\
HERMES & 37.5 & \textbf{39.8} & 71.6 & 49.7 & 55.8 & \textbf{54.3} & 55.1 & 70.4 & 50.0 & \underline{47.2} & 59.5\\
Expected Attention & \underline{42.7} & 34.5 & 69.5 & 48.9 & 56.9 & 52.1 & 54.5 & \textbf{70.9} & 51.2 & 45.0 & 59.5\\
TriAttention & \underline{42.7} & \underline{37.2} & 72.3 & \underline{50.8} & \underline{59.6} & 52.0 & \underline{55.8} & 69.5 & \underline{55.8} & 46.8 & \underline{60.4}\\
\midrule
\Ours{} & \textbf{44.4} & \textbf{39.8} & \underline{74.5} & \textbf{52.9} & \textbf{60.8} & \underline{53.7} & \textbf{57.2} & 69.0 & \textbf{57.8} & \underline{47.2} & \textbf{60.7}\\
\midrule
\multicolumn{12}{l}{\emph{Qwen3-VL-8B}}\\
Full KV & 55.6 & 51.3 & 80.1 & 62.4 & 69.2 & 61.1 & 65.2 & 67.6 & 66.0 & 56.9 & 64.5\\
\midrule
SWA & 29.0 & 31.9 & 58.9 & 39.9 & 50.0 & 50.7 & 50.3 & 66.8 & 45.6 & 40.6 & 54.8\\
InfiniPot-V & 44.0 & 35.4 & 64.5 & 48.0 & \underline{61.3} & \underline{56.1} & \underline{58.7} & \textbf{68.7} & 53.9 & 43.8 & 58.7\\
CoRDS & \underline{46.4} & \underline{40.7} & 63.1 & 50.1 & 56.8 & 51.9 & 54.3 & 67.3 & 56.3 & 43.8 & 58.7\\
KeyDiff & 36.3 & 36.3 & 63.8 & 45.5 & 54.0 & 52.3 & 53.2 & \underline{68.1} & 46.6 & 39.4 & 55.5\\
StreamMem & 39.5 & 35.4 & 63.1 & 46.0 & 55.4 & 54.6 & 55.0 & 67.3 & 53.2 & 41.7 & 57.3\\
HERMES & 35.9 & 31.9 & 64.5 & 44.1 & 56.3 & 54.2 & 55.3 & 67.6 & 52.9 & \underline{44.0} & 58.0\\
Expected Attention & 38.3 & 32.7 & 67.4 & 46.1 & 58.9 & \underline{56.1} & 57.5 & 67.0 & 55.3 & 43.8 & 58.3\\
TriAttention & \underline{46.4} & 37.2 & \underline{69.5} & \underline{51.0} & 60.7 & 55.4 & 58.1 & 67.3 & \underline{59.0} & \underline{44.0} & \underline{59.3}\\
\midrule
\Ours{} & \textbf{49.2} & \textbf{41.6} & \textbf{74.5} & \textbf{55.1} & \textbf{62.9} & \textbf{57.4} & \textbf{60.2} & 67.9 & \textbf{61.9} & \textbf{47.2} & \textbf{61.1}\\
\midrule
\multicolumn{12}{l}{\emph{LLaVA-OV-7B}}\\
Full KV & 43.1 & 23.0 & 76.6 & 47.6 & 54.7 & 46.6 & 50.6 & 66.2 & 53.4 & 49.3 & 58.8\\
\midrule
SWA & 33.5 & 21.2 & 61.0 & 38.6 & 49.0 & 43.0 & 46.0 & \underline{67.0} & 47.6 & 44.3 & 56.5\\
InfiniPot-V & 33.1 & 21.2 & 66.7 & 40.3 & 50.1 & \underline{45.8} & 47.9 & 65.9 & \underline{51.7} & \underline{45.4} & 57.2\\
CoRDS & 34.3 & 23.0 & 65.2 & 40.8 & 50.0 & 44.2 & 47.1 & 66.2 & 49.5 & 44.7 & 56.7\\
KeyDiff & 35.1 & 23.9 & 68.8 & 42.6 & 50.2 & 45.2 & 47.7 & \underline{67.0} & 51.0 & \textbf{45.6} & \underline{57.6}\\
StreamMem & 33.5 & \underline{24.8} & 68.8 & 42.3 & \textbf{51.9} & 44.3 & \underline{48.1} & \textbf{67.3} & 48.5 & 44.3 & 56.9\\
HERMES & 35.1 & 20.4 & 62.4 & 39.3 & 49.2 & 45.4 & 47.3 & \underline{67.0} & 49.0 & 44.3 & 56.9\\
Expected Attention & 31.9 & \textbf{26.5} & 66.7 & 41.7 & 50.7 & 44.8 & 47.7 & 66.5 & 50.7 & 44.3 & 57.0\\
TriAttention & \textbf{40.3} & 23.9 & \underline{70.2} & \textbf{44.8} & 51.2 & 44.8 & 48.0 & 65.9 & 51.2 & 45.0 & 57.0\\
\midrule
\Ours{} & \underline{37.1} & 23.9 & \textbf{70.9} & \underline{44.0} & \underline{51.8} & \textbf{46.0} & \textbf{48.9} & 66.5 & \textbf{52.9} & \textbf{45.6} & \textbf{57.9}\\
\bottomrule
\end{tabular}
\caption{\textbf{Task-level accuracy at the 3K budget.} The numbers of Table~\ref{tab:v2_main} split by task group (MLVU), video duration (Video-MME medium and long videos) and video length (LongVideoBench: videos of at most 60, 600 and 3600 seconds; the 15 and 60 second groups are merged), in \%; Avg is the value of the main table, the macro average (unweighted mean) of the groups of each benchmark, which for LongVideoBench are the four length groups of 15, 60, 600 and 3600 seconds. Bold: best, underlined: second best per column and backbone.}
\label{tab:v2_main_ext_3k}
\end{table*}

\begin{table*}[!t]
\centering
\normalsize
\setlength{\tabcolsep}{4pt}
\renewcommand{\arraystretch}{0.95}
\begin{tabular}{l cccc ccc cccc}
\toprule
 & \multicolumn{4}{c}{MLVU} & \multicolumn{3}{c}{Video-MME} & \multicolumn{4}{c}{LongVideoBench}\\
\cmidrule(lr){2-5}\cmidrule(lr){6-8}\cmidrule(lr){9-12}
Method & Single & Multi & Holistic & Avg & Medium & Long & Avg & $\leq$60\,s & 600\,s & 3600\,s & Avg\\
\midrule
\multicolumn{12}{l}{\emph{Qwen2.5-VL-7B}}\\
Full KV & 48.0 & 38.1 & 75.9 & 54.0 & 64.7 & 54.2 & 59.4 & 69.8 & 63.6 & 52.3 & 63.9\\
\midrule
SWA & 29.8 & 30.1 & 62.4 & 40.8 & 48.0 & 46.0 & 47.0 & 69.5 & 47.1 & 43.3 & 57.3\\
InfiniPot-V & 34.3 & 29.2 & 66.7 & 43.4 & 55.1 & 49.4 & 52.3 & 69.5 & 49.8 & 44.3 & 58.2\\
CoRDS & 28.6 & 31.9 & 66.7 & 42.4 & 51.3 & 47.4 & 49.4 & 69.8 & 49.0 & 43.3 & 57.9\\
KeyDiff & 27.0 & \underline{32.7} & 67.4 & 42.4 & 53.2 & 48.4 & 50.8 & 69.3 & 50.2 & 45.2 & 58.4\\
StreamMem & 33.1 & 31.0 & 68.8 & 44.3 & 51.2 & 48.4 & 49.8 & \underline{70.6} & 48.8 & 43.4 & 58.3\\
HERMES & 34.7 & 27.4 & 66.7 & 42.9 & 53.7 & 49.9 & 51.8 & \textbf{70.9} & 46.6 & \textbf{46.5} & 58.7\\
Expected Attention & 35.1 & 30.1 & \underline{69.5} & 44.9 & 53.2 & 50.0 & 51.6 & 69.3 & 47.6 & \underline{46.3} & 58.0\\
TriAttention & \underline{37.1} & \textbf{38.1} & \textbf{73.0} & \underline{49.4} & \underline{56.3} & \underline{51.1} & \underline{53.7} & 69.0 & \textbf{53.2} & \textbf{46.5} & \underline{59.3}\\
\midrule
\Ours{} & \textbf{39.5} & \textbf{38.1} & \textbf{73.0} & \textbf{50.2} & \textbf{56.9} & \textbf{51.9} & \textbf{54.4} & 69.3 & \underline{52.9} & \textbf{46.5} & \textbf{59.4}\\
\midrule
\multicolumn{12}{l}{\emph{Qwen3-VL-8B}}\\
Full KV & 55.6 & 51.3 & 80.1 & 62.4 & 69.2 & 61.1 & 65.2 & 67.6 & 66.0 & 56.9 & 64.5\\
\midrule
SWA & 25.4 & 32.7 & 56.0 & 38.1 & 48.0 & 47.0 & 47.5 & 64.5 & 42.5 & 36.9 & 52.0\\
InfiniPot-V & 33.9 & 34.5 & 61.0 & 43.1 & 54.6 & 51.7 & 53.1 & \textbf{66.8} & 45.4 & 40.8 & 54.8\\
CoRDS & \underline{36.3} & 29.2 & \underline{63.8} & 43.1 & 50.2 & 48.9 & 49.6 & 65.9 & 43.7 & 37.4 & 53.1\\
KeyDiff & 27.8 & 32.7 & 57.4 & 39.3 & 48.3 & 46.9 & 47.6 & 62.3 & 39.6 & 35.1 & 49.6\\
StreamMem & 31.5 & 31.0 & 56.7 & 39.7 & 47.8 & 50.3 & 49.1 & 63.4 & 49.8 & 41.3 & 54.3\\
HERMES & 33.5 & 32.7 & 57.4 & 41.2 & 49.6 & 50.6 & 50.1 & \underline{66.2} & 47.8 & \underline{42.0} & \underline{55.5}\\
Expected Attention & 33.9 & 32.7 & 63.1 & 43.2 & \underline{54.7} & \underline{53.2} & \underline{53.9} & 65.1 & 48.8 & 41.0 & 54.9\\
TriAttention & 35.1 & \textbf{37.2} & \textbf{66.7} & \underline{46.3} & 54.3 & 52.9 & 53.6 & 64.3 & \underline{52.7} & 39.7 & 55.1\\
\midrule
\Ours{} & \textbf{40.3} & \underline{36.3} & \textbf{66.7} & \textbf{47.8} & \textbf{59.9} & \textbf{55.9} & \textbf{57.9} & \underline{66.2} & \textbf{59.2} & \textbf{42.9} & \textbf{58.6}\\
\midrule
\multicolumn{12}{l}{\emph{LLaVA-OV-7B}}\\
Full KV & 43.1 & 23.0 & 76.6 & 47.6 & 54.7 & 46.6 & 50.6 & 66.2 & 53.4 & 49.3 & 58.8\\
\midrule
SWA & 30.6 & 18.6 & 61.0 & 36.7 & 47.0 & 43.0 & 45.0 & 66.5 & 49.0 & 42.2 & 56.0\\
InfiniPot-V & 31.9 & 19.5 & 64.5 & 38.6 & 48.2 & \underline{45.1} & 46.7 & \textbf{67.9} & \underline{49.5} & 43.3 & \textbf{57.2}\\
CoRDS & 31.5 & 19.5 & 61.0 & 37.3 & 47.4 & 42.9 & 45.2 & 65.9 & 46.8 & 43.3 & 55.5\\
KeyDiff & 31.9 & 22.1 & \underline{66.0} & 40.0 & 48.1 & 44.1 & 46.1 & 66.8 & 46.1 & 43.1 & 55.7\\
StreamMem & 30.2 & 22.1 & 63.1 & 38.5 & \textbf{51.4} & 43.0 & \underline{47.2} & \underline{67.6} & 45.6 & 42.4 & 55.8\\
HERMES & 31.0 & 20.4 & 59.6 & 37.0 & 46.9 & 43.0 & 44.9 & 66.2 & 48.8 & \underline{43.8} & \underline{56.2}\\
Expected Attention & \underline{34.3} & 20.4 & 61.0 & 38.5 & 48.7 & 44.3 & 46.5 & 65.7 & 46.6 & 41.5 & 54.9\\
TriAttention & \underline{34.3} & \underline{23.9} & \textbf{66.7} & \underline{41.6} & 50.2 & 43.3 & 46.8 & 65.4 & 48.3 & \underline{43.8} & 55.7\\
\midrule
\Ours{} & \textbf{34.7} & \textbf{24.8} & \underline{66.0} & \textbf{41.8} & \underline{50.7} & \textbf{45.2} & \textbf{47.9} & 65.4 & \textbf{52.9} & \textbf{45.2} & \textbf{57.2}\\
\bottomrule
\end{tabular}
\caption{\textbf{Task-level accuracy at the 1K budget.} The numbers of Table~\ref{tab:v2_main} split by task group (MLVU), video duration (Video-MME medium and long videos) and video length (LongVideoBench: videos of at most 60, 600 and 3600 seconds; the 15 and 60 second groups are merged), in \%; Avg is the value of the main table, the macro average (unweighted mean) of the groups of each benchmark, which for LongVideoBench are the four length groups of 15, 60, 600 and 3600 seconds. Bold: best, underlined: second best per column and backbone.}
\label{tab:v2_main_ext_1k}
\end{table*}

\subsection{Per-Category Results}
\label{app:v2_per_category}

\paragraph{Long-video benchmarks.}
Tables~\ref{tab:v2_main_ext_6k}, \ref{tab:v2_main_ext_3k} and~\ref{tab:v2_main_ext_1k} split the accuracies of Table~\ref{tab:v2_main} by task group (MLVU), video duration (Video-MME medium and long videos) and video length (LongVideoBench), at each budget.

\clearpage
\begin{table*}[!t]
\centering
\normalsize
\setlength{\tabcolsep}{5pt}
\renewcommand{\arraystretch}{0.95}
\begin{tabular}{lcccccc}
\toprule
\textbf{Method} & Scene sum. & Action cap. & Event order & Order (Y/N) & Happened (Y/N) & All\\
\midrule
\multicolumn{7}{l}{\emph{Qwen2.5-VL-7B}}\\
\midrule
SWA & 59.4 & 49.8 & \underline{33.2} & 48.2 & 70.9 & 52.9\\
InfiniPot-V & 59.4 & 53.2 & 33.1 & 49.7 & \underline{72.1} & 54.0\\
CoRDS & 59.3 & 54.8 & \textbf{33.9} & 49.2 & \textbf{72.5} & \underline{54.4}\\
KeyDiff & 59.3 & 54.0 & 29.8 & \underline{51.1} & 70.5 & 53.5\\
StreamMem & 58.3 & 54.2 & 28.3 & 48.8 & 68.8 & 52.2\\
HERMES & \underline{59.8} & \underline{55.7} & 29.8 & 48.8 & 69.1 & 53.1\\
Expected Attention & \underline{59.8} & 54.5 & 32.4 & 50.4 & 68.1 & 53.6\\
TriAttention & 59.0 & 54.2 & 28.5 & \textbf{52.1} & 68.5 & 53.1\\
\midrule
\Ours{} & \textbf{62.1} & \textbf{59.2} & 30.8 & 49.7 & 70.1 & \textbf{54.9}\\
\midrule
\multicolumn{7}{l}{\emph{Qwen3-VL-8B}}\\
\midrule
SWA & 62.5 & 58.8 & \textbf{35.2} & 61.7 & 70.6 & 58.4\\
InfiniPot-V & 65.9 & 57.3 & 32.1 & 61.9 & 72.1 & 58.6\\
CoRDS & 65.5 & 59.0 & 31.6 & \underline{62.8} & 72.3 & 59.0\\
KeyDiff & 63.8 & 56.0 & 29.8 & 58.7 & \textbf{73.2} & 57.1\\
StreamMem & 64.2 & 58.5 & 34.2 & 62.2 & 69.8 & 58.4\\
HERMES & 66.1 & 60.7 & 34.4 & 60.8 & 70.5 & \underline{59.1}\\
Expected Attention & 64.0 & 58.0 & \underline{34.5} & 61.0 & \underline{72.5} & 58.7\\
TriAttention & \underline{66.8} & \underline{60.8} & 34.0 & \textbf{63.5} & 70.1 & \textbf{59.7}\\
\midrule
\Ours{} & \textbf{68.5} & \textbf{63.5} & 31.9 & 61.6 & 69.6 & \textbf{59.7}\\
\midrule
\multicolumn{7}{l}{\emph{LLaVA-OV-7B}}\\
\midrule
SWA & \underline{50.4} & 41.2 & 16.4 & \underline{66.7} & 69.1 & 49.9\\
InfiniPot-V & 48.5 & 41.0 & 17.1 & 65.6 & 69.2 & 49.4\\
CoRDS & 48.2 & 42.8 & 16.4 & 65.7 & \textbf{71.3} & 50.0\\
KeyDiff & 48.6 & 42.5 & 17.6 & 65.1 & 69.2 & 49.7\\
StreamMem & 49.2 & \underline{43.3} & 16.9 & 65.5 & 69.9 & \underline{50.1}\\
HERMES & 49.3 & 40.7 & \underline{18.2} & 65.5 & 69.6 & 49.8\\
Expected Attention & 47.7 & 40.0 & 16.9 & \textbf{67.0} & \underline{70.2} & 49.5\\
TriAttention & 49.0 & 41.3 & 16.9 & 66.2 & 67.6 & 49.3\\
\midrule
\Ours{} & \textbf{55.0} & \textbf{48.8} & \textbf{18.9} & \textbf{67.0} & 68.2 & \textbf{52.6}\\
\bottomrule
\end{tabular}
\caption{\textbf{VStream-QA accuracy by question type.} The runs of Table~\ref{tab:v2_streaming} (about 6K visual tokens per head), split by the type of the question, over the ego-centric (RVS-Ego) and movie (RVS-Movie) splits together; the number of questions is 734 (Scene sum.), 600 (Action cap.), 614 (Event order), 724 (Order (Y/N)), 698 (Happened (Y/N)). Scene sum.: scene summary, Action cap.: action caption, Event order: what is the order of the events, Order (Y/N): judging whether an order is correct, Happened (Y/N): whether something happened. Open-ended answers are judged by \texttt{gpt-5-mini}. \textbf{Bold}: best, \underline{underline}: second best per column and backbone.}
\label{tab:v2_streaming_pertask}
\end{table*}

\paragraph{Streaming benchmark.}
Table~\ref{tab:v2_streaming_pertask} splits the accuracies of Table~\ref{tab:v2_streaming} by question type, over the ego-centric (RVS-Ego) and movie (RVS-Movie) splits together.

\begin{table*}[!t]
\centering
\small
\setlength{\tabcolsep}{4pt}
\begin{tabular}{l ccc ccc ccc}
\toprule
 & \multicolumn{3}{c}{Qwen2.5-VL} & \multicolumn{3}{c}{Qwen3-VL} & \multicolumn{3}{c}{LLaVA-OV}\\
\cmidrule(lr){2-4}\cmidrule(lr){5-7}\cmidrule(lr){8-10}
Query & 6K & 3K & 1K & 6K & 3K & 1K & 6K & 3K & 1K\\
\midrule
Recent visual tokens & 49.4 & 47.0 & 44.8 & 49.2 & 45.5 & 43.5 & 43.6 & 42.6 & 38.5\\
Q-Filters (vision) & 49.2 & 48.3 & 44.4 & \underline{54.5} & \underline{52.2} & 44.6 & 44.1 & 41.4 & 39.6\\
Chat template & 49.0 & 43.8 & 44.3 & 53.5 & 47.7 & 40.7 & 42.6 & 39.8 & 38.2\\
\textsc{describe} & 51.1 & 47.7 & 43.9 & 53.3 & 48.7 & 42.7 & 42.2 & 40.7 & 37.0\\
\textsc{explain} & 49.4 & 44.6 & 43.5 & 52.5 & 45.7 & 38.8 & 41.6 & 42.3 & 37.2\\
\textsc{summarize} & 49.6 & 46.8 & 45.9 & 50.8 & 46.6 & 41.8 & 42.8 & 41.2 & 39.2\\
Expected Attention & 50.3 & 47.9 & 45.1 & 52.6 & 48.1 & 42.2 & 42.5 & 42.3 & 38.8\\
TriAttention & \underline{53.2} & \underline{50.8} & \underline{49.4} & 54.2 & 51.0 & \underline{46.3} & \underline{45.2} & \textbf{44.8} & \underline{41.6}\\
\midrule
\Ours{} & \textbf{55.9} & \textbf{52.9} & \textbf{50.2} & \textbf{58.1} & \textbf{55.1} & \textbf{47.8} & \textbf{45.9} & \underline{44.0} & \textbf{41.8}\\
\bottomrule
\end{tabular}
\caption{\textbf{Accuracy of each query at the three budgets.} MLVU accuracy (\%, average over task groups) when the cache is compressed with each query, at about 6K / 3K / 1K visual tokens per head, with the selector, the streaming protocol and the budgets of Supp.~\ref{app:v2_protocol} held fixed. Expected Attention uses the mean and covariance of the visual-token queries of the newest chunk, without the value norm, so that only the query differs; TriAttention is the row of Table~\ref{tab:v2_main}, with its full score. Bold: best, underlined: second best per column.}
\label{tab:v2_query_ext}
\end{table*}

\subsection{Query Analysis Across Budgets}
\label{app:v2_query_budgets}

The query analysis of Sec.~\ref{sec:v2_analysis} compares the queries at the 6K budget; Table~\ref{tab:v2_query_ext} gives the MLVU accuracy of every one of them at the 6K, 3K and 1K budgets, with the selector and the streaming protocol of Supp.~\ref{app:v2_protocol} held fixed (the queries are described in Supp.~\ref{app:v2_query_cfg}).
Expected Attention is the query-only version of Supp.~\ref{app:v2_query_cfg}; the TriAttention row is copied from Table~\ref{tab:v2_main}, with its full score.
Q-Filters, which induces no attention distribution and is therefore not among the queries of Fig.~\ref{fig:v2_gap} (Supp.~\ref{app:v2_query_cfg}), is less accurate than \Ours{} in all nine settings.

\section{Additional Qualitative Results}
\label{app:qualitative}

\begin{figure*}[t]
\centering
\includegraphics[width=\textwidth]{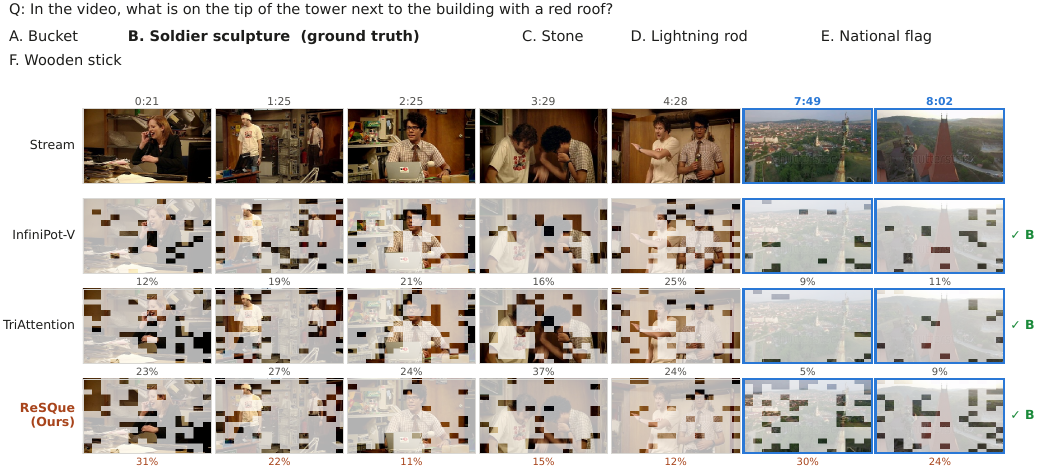}
\caption{\textbf{Qualitative example} (MLVU, LLaVA-OV, 6K tokens per head). Patches each method keeps in one head (layer 16, KV head 0) when the question arrives, with evicted patches faded and the share kept under each frame. The blue frames lie in the evidence clip (7:45--8:07), which receives 51\% of the question's attention on the uncompressed cache but spans only 5.0\% of the frames. \Ours keeps $30\%$ of the patches of the evidence frames, against $10\%$ and $8\%$; all three methods answer correctly.}
\label{fig:qual_extra_llava}
\end{figure*}

\begin{figure*}[t]
\centering
\includegraphics[width=\textwidth]{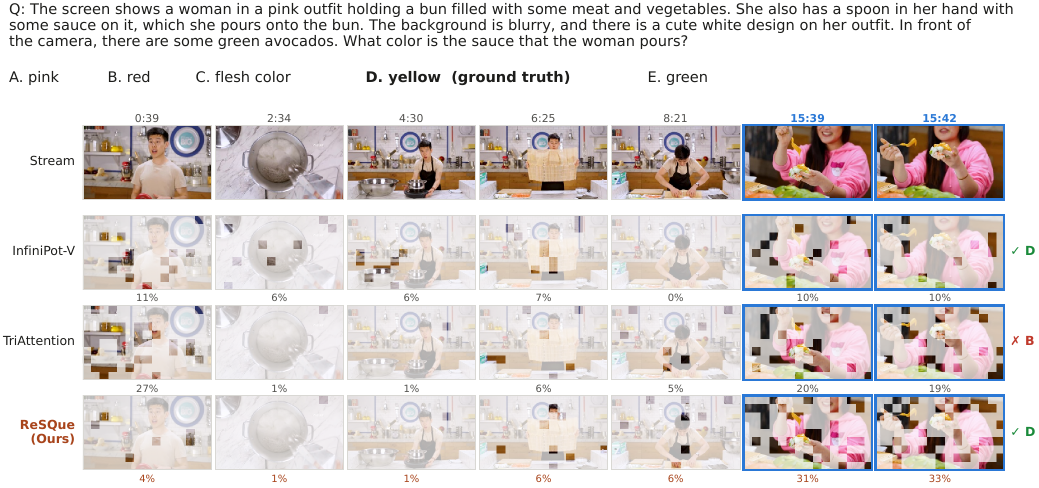}
\caption{\textbf{Qualitative example} (LongVideoBench, Qwen3-VL, 6K tokens per head). Patches each method keeps in one head (layer 23, KV head 2) when the question arrives, with evicted patches faded and the share kept under each frame. The blue frames lie at the moment the question refers to (about 15:40), which receives 14\% of the question's attention on the uncompressed cache but spans only 0.5\% of the frames. \Ours keeps $32\%$ of their patches, against $10\%$ for InfiniPot-V and $19\%$ for TriAttention; InfiniPot-V and \Ours answer correctly, TriAttention does not.}
\label{fig:qual_extra_q3}
\end{figure*}

\begin{figure*}[t]
\centering
\includegraphics[width=\textwidth]{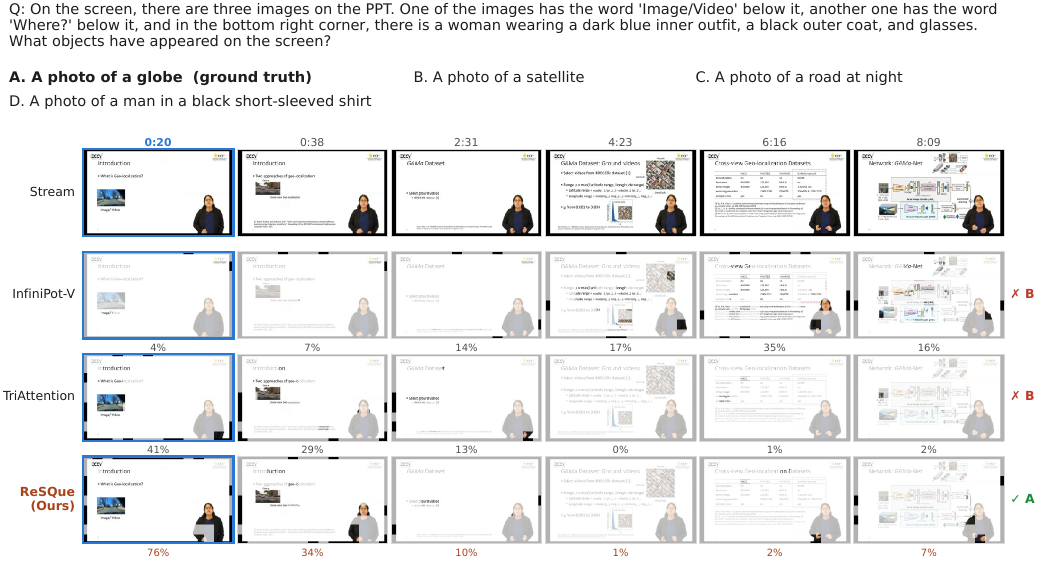}
\caption{\textbf{Qualitative example} (LongVideoBench, Qwen2.5-VL, 6K tokens per head). Patches each method keeps in one head (layer 14, KV head 0) when the question arrives, with evicted patches faded and the share kept under each frame. The blue frame is the moment the question refers to (about 0:20), which receives 6\% of the question's attention on the uncompressed cache but spans only 0.3\% of the frames. \Ours keeps $76\%$ of its patches in this head, against $4\%$ and $41\%$ ($37\%$ averaged over all heads, against $9\%$ and $15\%$), and is the only one to answer correctly.}
\label{fig:qual_extra_lvb25}
\end{figure*}

Fig.~\ref{fig:qual_extra_llava} to Fig.~\ref{fig:qual_extra_lvb25} show three more questions in the format of Fig.~\ref{fig:qual}: one from MLVU, whose evidence clips are labelled, and two from LongVideoBench, whose evidence is the annotated moment the question refers to.
In each, we draw the head whose question attention on the uncompressed cache is the most concentrated on the evidence, and compare what InfiniPot-V, TriAttention and \Ours{} hold in it when the question arrives (6K tokens per head).

\end{document}